\documentclass[10pt,twocolumn,letterpaper]{article}

\usepackage{cvpr}              
\newcommand{\myparagraph}[1]{\vspace{3pt}\noindent\textbf{#1}}

\definecolor{cvprblue}{rgb}{0.21,0.49,0.74}
\usepackage[pagebackref,breaklinks,colorlinks,allcolors=cvprblue]{hyperref}
\usepackage{color}
\usepackage{hyperref}
\usepackage{graphicx}
\usepackage{multirow}
\usepackage{xspace}
\usepackage{graphicx}
\usepackage{booktabs}
\usepackage{multirow}
\usepackage{pifont}
\usepackage{tabularx}
\usepackage{colortbl}
\usepackage[symbol]{footmisc}
\usepackage{wrapfig}
\usepackage{tcolorbox}
\usepackage[title]{appendix}

\newcommand{\xmark}{\ding{55}}
\definecolor{checkmark}{HTML}{40826D}
\definecolor{xmark}{HTML}{E62020}
\newcommand{\ccheck}{{\textcolor{checkmark}{\large\checkmark}}}
\newcommand{\ccross}{{\textcolor{xmark}{\large\xmark}}}

\newcommand{\bccheck}{\large\checkmark}
\newcommand{\bccross}{\large\xmark}

\newcommand{\ccb}{\cellcolor{blue!10}}

\newcommand{\ccdb}{\cellcolor{blue!20}}
\newcommand{\ccnb}{\cellcolor{blue!30}}
\newcommand{\cclc}{\cellcolor{cyan!15}}
\newcommand{\ccdc}{\cellcolor{cyan!30}}
\definecolor{green}{HTML}{B9DCB7} 
\definecolor{pink}{HTML}{fad0c3}

\usepackage{soul} 

\def\confName{CVPR}
\def\confYear{2025}

\title{Distilling Multi-modal Large Language Models for Autonomous Driving}

\author{Deepti Hegde$^{1\dagger\ast}$,
Rajeev Yasarla$^{2\ast}$,
Hong Cai$^{2}$ , 
Shizhong Han$^{2}$,
Apratim Bhattacharyya$^{2}$, \\
Shweta Mahajan$^{2}$,
Litian Liu$^{2}$,
Risheek Garrepalli$^{2}$,
Vishal M. Patel$^{1}$,
Fatih Porikli$^{2}$ \\
[2mm]
$^1$~Johns Hopkins University,
$^2$~Qualcomm AI Research$^{\ddagger}$\\
\normalsize{
$^\ast$Equal contribution}
}

\newcommand{\ours}{{DiMA}\xspace}
\newcommand{\oursp}{{DiMA+}\xspace}

\newcommand\blfootnote[1]{%
  \begingroup
  \renewcommand\thefootnote{}\footnote{#1}%
  \addtocounter{footnote}{-1}%
  \endgroup
}

\begin{document}
\maketitle

\setlength{\abovedisplayskip}{3pt}
\setlength{\belowdisplayskip}{3pt}
\begin{abstract}
     Autonomous driving demands safe motion planning, especially in critical ``long-tail'' scenarios. Recent end-to-end autonomous driving systems leverage large language models (LLMs) as planners to improve generalizability to rare events. However, using LLMs at test time introduces high computational costs. To address this, we propose \ours, an end-to-end autonomous driving system that maintains the efficiency of an LLM-free (or vision-based) planner while leveraging the world knowledge of an LLM. \ours distills the information from a multi-modal LLM to a vision-based end-to-end planner through a set of specially designed surrogate tasks. Under a joint training strategy, a scene encoder common to both networks produces structured representations that are semantically grounded as well as aligned to the final planning objective. Notably, the LLM is optional at inference, enabling robust planning without compromising on efficiency. Training with \ours results in a $37\%$ reduction in the L2 trajectory error and an $80\%$ reduction in the collision rate of the vision-based planner, as well as a $44\%$ trajectory error reduction in long-tail scenarios. \ours also achieves state-of-the-art performance on the nuScenes planning benchmark.

\end{abstract}

\begin{figure}
    \centering
    \includegraphics[width=\linewidth]{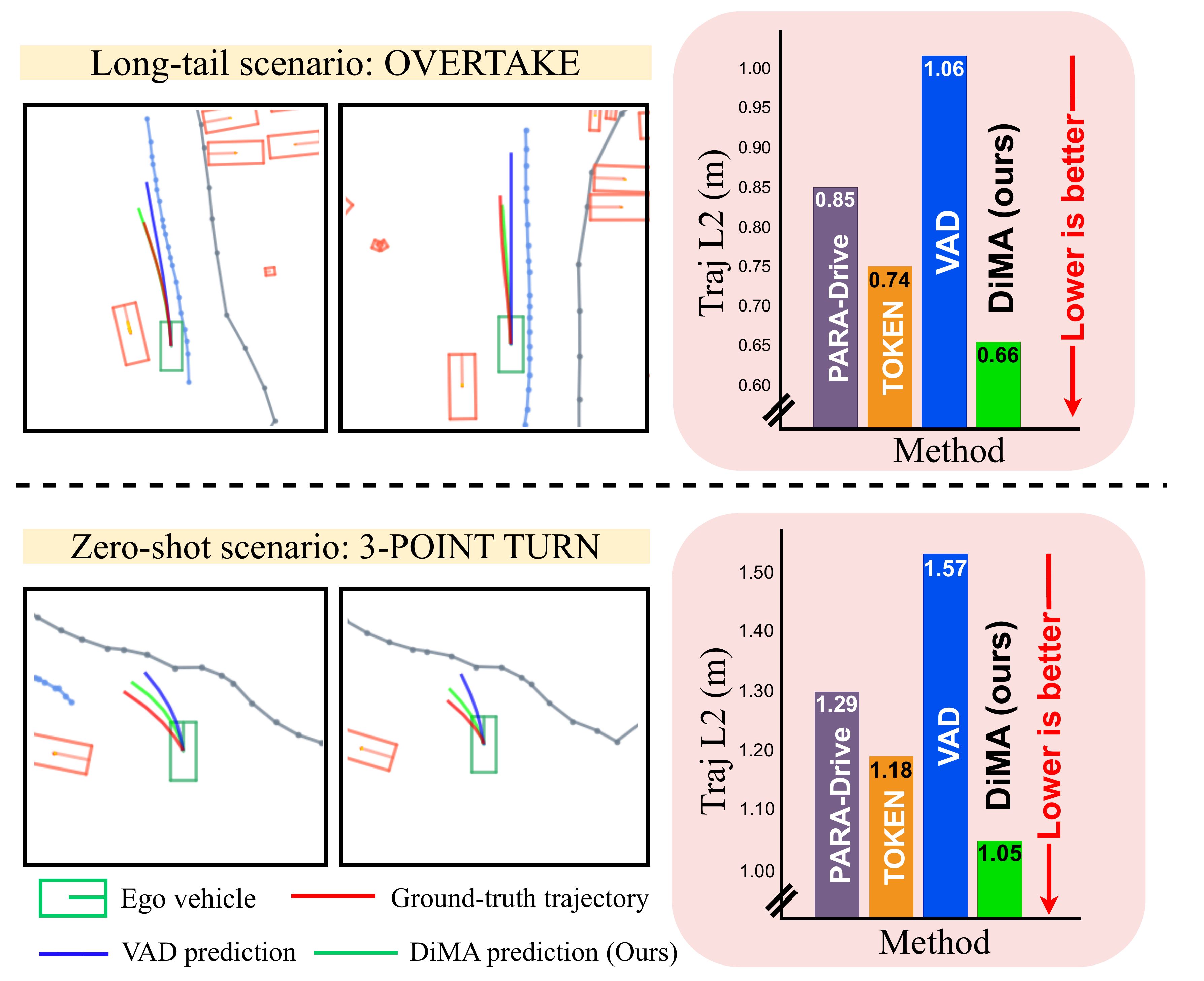}
     \caption{Comparison of planning performance in long-tail scenarios from nuScenes: \ours-VAD demonstrates greater robustness compared to {VAD}~\cite{jiang2023vad} in long-tail navigation scenarios such as overtaking a vehicle and performing a 3-point turn. \ours-VAD also outperforms recent vision-planner PARA-Drive \cite{weng2024drive} and LLM planner TOKEN \cite{tian2024tokenize}. Notably, the 3-point turn is a zero-shot scenario that is only present in the validation set.}
     \vspace{-5pt}
    \label{fig:visualization}
\end{figure} 
\noindent
\blfootnote{$\dagger$Work done while an intern at Qualcomm AI Research.}
\blfootnote{${\ddagger}$Qualcomm AI Research is an initiative of Qualcomm Technologies, Inc.}
\vspace{-10pt}
\section{Introduction}
\label{sec:intro}

Research in autonomous driving has moved away from the traditional strategy of integrating sequential, independently trained models \cite{li2022bevformer,yin2021center,philion2020lift,wang2024mv2dfusion,chai2019multipath,liu2021multimodal,ngiam2021scene}, and moved towards developing multi-task systems that are trained in an end-to-end manner \cite{jiang2023vad,hu2022st,chen2022learning,casas2021mp3,chitta2021neat,wu2022trajectory}. These approaches demonstrate improved performance and interpretability while being highly efficient. However, they struggle with long-tail navigation and perception scenarios, primarily because they are trained on task-specific, limited datasets \cite{caesar2020nuscenes,dosovitskiy2017carla,ettinger2021large}. 

Large language models (LLMs) have emerged as a promising solution to this issue. Trained on vast, internet-scale datasets, LLMs can leverage world knowledge to generalize to unseen or rare scenarios. These models can perform high-level reasoning tasks using mechanisms such as chain-of-thought \cite{wei2022chain}. Going beyond text-based prompting, multi-modal large language models (MLLMs) integrate image and video inputs to the LLM, enabling tasks such as visual question answering and dense captioning \cite{li2023blip,liu2024visual}. Recent end-to-end autonomous driving systems leverage LLMs to achieve higher interpretability through language-guided reasoning and superior robustness to long-tail scenarios \cite{sima2023drivelm,tian2024drivevlm,wang2023drivemlm,tian2024tokenize,hwang2024emma,wang2024omnidrive}. To differentiate between these methods, we call end-to-end planners that depend on LLMs to perform trajectory prediction as ``LLM-based'' planners and ones that do not as ``vision-based'' planners \cite{jiang2023vad,hu2023planning,weng2024drive}. In spite of their recent success ``LLM-based'' planners face significant challenges.

Firstly, LLM-based planners require a significant amount of computational overhead at test time, limiting their practicality.
This work addresses a core question: \textit{how can we harness LLMs' world knowledge while preserving the efficiency of vision-based planners?} Secondly, bridging the visual and language domains is inherently more complex for end-to-end planning tasks compared to general MLLM tasks \cite{li2023blip,liu2024visual}, with the additional hurdle of limited training data.
Standard image tokenization strategies rely on frozen pre-trained image encoders \cite{radford2021learning} to generate dense, unstructured multi-modal token embeddings. \cite{li2023blip,sima2023drivelm,liu2024visual}. We suggest that MLLMs meant for end-to-end autonomous driving benefit from structured inputs that explicitly model scene components. Furthermore, we suggest that updating these features during training results in richer representations.

\begin{figure*}[ht!]
    \centering
    \includegraphics[width=0.9\linewidth]{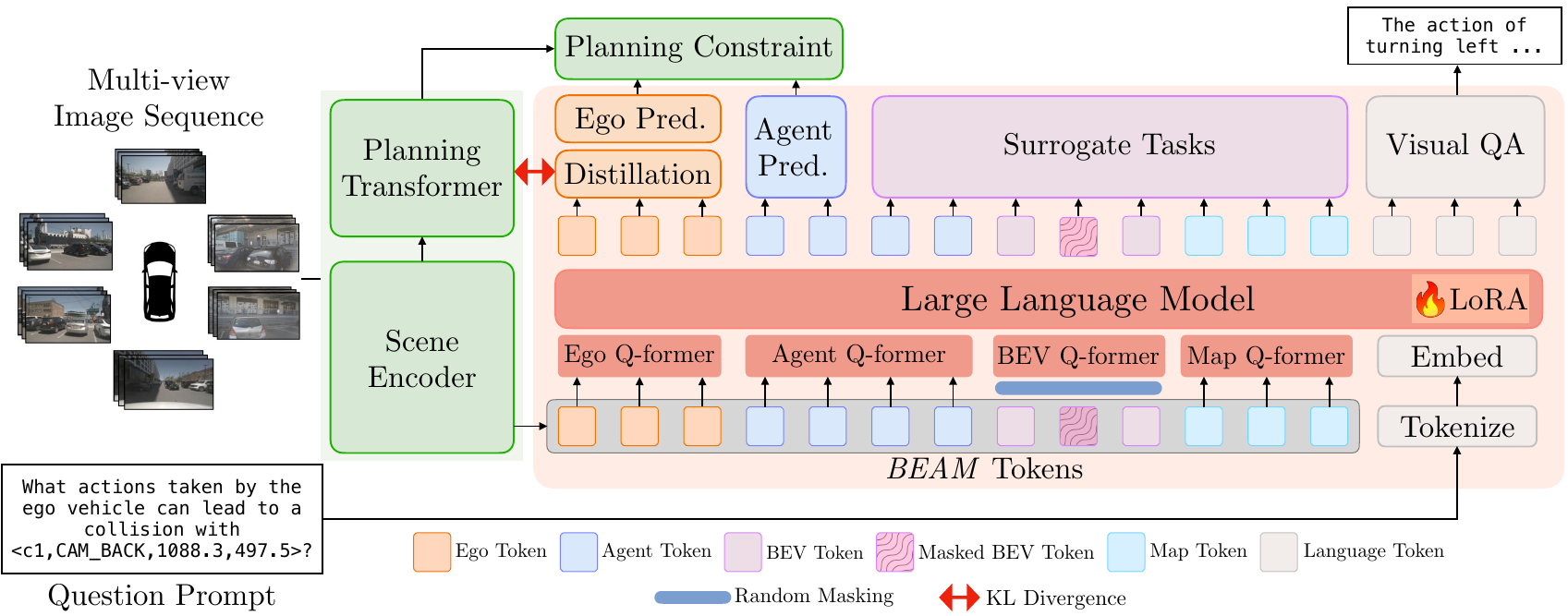}
    \vspace{-5pt}
    \caption{Overview of \ours. The input to the framework is a multi-view image sequence and a question text prompt. The \sethlcolor{green}\hl{vision-based end-to-end planner} consists of a scene encoder and a planning transformer. The scene encoder learns structured latent representations in the form of \textbf{b}ird's-eye-view, \textbf{e}go, \textbf{a}gent, and \textbf{m}ap ($BEAM$) token embeddings and acts as a trainable tokenizer for the \sethlcolor{pink}\hl{multi-modal large language model} (MLLM). The planning transformer is trained under standard planning constraints \cite{jiang2023vad,hu2023planning}. The MLLM is trained for planning, visual question answering, distillation, and a series of surrogate tasks.}
    \label{fig:main}
    \vspace{-5pt}
\end{figure*}

To address these challenges, we propose \textbf{\ours}, a novel framework for \underline{\textbf{Di}}stilling \textbf{M}ulti-modal Large Language Models for \underline{\textbf{A}}utonomous driving. We introduce a joint-training scheme between a vision-based planner and an MLLM that enables the learning of robust, grounded and disentangled scene representations that are aligned to the final objective of planning. Specifically, we use the vision-based planner as a tokenizer, and pass the learned scene representations to the MLLM, providing a more structured input. The MLLM is then trained for the tasks of visual question-answering, trajectory estimation, and a series of surrogate tasks designed to ground the multi-modal inputs in language. The vision-based planner is simultaneously trained for trajectory estimation, while also distilling features from the MLLM planning head to the planning transformer. Crucially, \textbf{the MLLM can be discarded when performing planning inference}, maintaining the efficiency of the vision-based planner while utilizing the knowledge of the language model. Optionally, the MLLM can support language-guided reasoning through visual question answering. Our experiments show that DiMA is more robust to challenging long-tail scenarios compared to baselines and state-of-the-art methods \cite{weng2024drive,tian2024tokenize}.  Figure~\ref{fig:visualization} confirms this both quantitatively and qualitatively. We list the contributions of the work below:

\begin{enumerate}
    \item We introduce \textbf{\ours}, an end-to-end autonomous driving framework that distills knowledge from an MLLM to a vision-based planner to ensure robustness to long-tail events while maintaining efficiency. \ours is capable of planning as well as visual question-answering.
    \item We propose a distillation task along with the following surrogate tasks to align the objectives of the vision-based planner and the MLLM: (i) masked token reconstruction (ii) future token prediction (iii) scene editing.
    \item  \ours outperforms both vision-based and LLM end-to-end planners, achieving state-of-the-art results on the nuScenes planning benchmark. Training with \ours results in a $37\%$ reduction in the L2 trajectory error and an $80\%$ reduction in the collision rate of the vision-based planner, as well as a $44\%$ trajectory error reduction in long-tail scenarios.
\end{enumerate}


\newcommand{\lliu}[1]{{\color{brown}{#1}}}

\section{Related Work}
\subsection{End-to-End Autonomous Driving}

Autonomous driving (AD) systems are generally modular, consisting of tasks such as perception \cite{li2022bevformer,yin2021center,philion2020lift,wang2024mv2dfusion}, motion prediction \cite{chai2019multipath,liu2021multimodal,ngiam2021scene}, and planning \cite{codevilla2019exploring}.
Classic AD systems use standalone models for each task where each module is optimized separately and combined sequentially.
Alternatively, some approaches incorporate these modular tasks into a multi-task learning paradigm, where tasks shared a common feature extraction process but have task-specific heads \cite{chitta2022transfuser,zhang2022beverse, zeng2019end, liang2022effective}.
Recently, research has moved toward optimizing autonomous driving systems in an end-to-end manner, aiming to directly predict future trajectories \cite{jiang2023vad,hu2022st,chen2022learning,casas2021mp3,chitta2021neat,wu2022trajectory}.
Along this line of work, UniAD \cite{hu2023planning} introduces a query-based design that integrates perception and prediction components, enabling an end-to-end planning scheme. 
VAD \cite{jiang2023vad} utilizes a vectorized scene representation, replacing the dense rasterized representations from \cite{hu2023planning}, reducing computational cost and improving planning performance. 
While effective in general navigation scenarios, these methods struggle in difficult long-tail events (see Figure \ref{fig:visualization}). In this work, we propose a method for end-to-end autonomous driving that is robust to such scenarios.

\subsection{LLMs for Autonomous Driving} 

The recent success of large language models \cite{touvron2023llama, touvron2023llama2, dubey2024llama} demonstrates their ability to process complex contextual information with logical reasoning and to communicate in a human-interpretable manner, attracting significant attention in autonomous driving \cite{shao2024lmdrive, sima2023drivelm, tian2024drivevlm, mao2023gpt, sha2023languagempc, xu2024drivegpt4, chen2024driving, ding2023hilm, fu2024drive, liu2023mtd, wang2023drivemlm, wendilu}. These works investigate the capabilities of language models to generalize to novel scenarios as well as their ability to reason about the scene in the form of text.

\noindent\textbf{Prompting LLMs.} One category of approaches involves prompting LLMs with text using reasoning frameworks such as chain-of-thought \cite{wei2022chain}. LanguageMPC \cite{sha2023languagempc} leverages the common-sense reasoning capacity of LLM to make high-level driving decisions, which are then translated into bottom-level control signals.
GPT-Driver \cite{mao2023gpt} converts observations and ego-states into language prompts, framing driving planning as a text-based task for LLMs.

\noindent\textbf{End-to-end Driving with MLLMs.} Another approach is to integrate multi-modal LLMs into end-to-end train frameworks for autonomous vehicles. LMDrive \cite{shao2024lmdrive} integrates a vision encoder with a large language model, enabling natural language instruction in autonomous driving. 
DriveLM introduces multi-step question answering, going beyond single-round interactions \cite{malla2023drama, qian2024nuscenes, sachdeva2024rank2tell}.
DriveVLM \cite{tian2024drivevlm} leverages MLLMs and a chain-of-thought process to enhance scene understanding and planning.

We point out a challenge in the current MLLM strategy for autonomous driving. We suggest that in order to reason about vehicle dynamics, the language model benefits from a structured input that explicitly models scene components instead of dense, unstructured input to the MLLM as in \cite{wang2024omnidrive,shao2024lmdrive,tian2024drivevlm,wang2023drivemlm}. In TOKEN \cite{tian2024tokenize}, the scene encoder from \cite{weng2024drive} is used as a tokenizer. However, it is frozen during training. In our approach, the input to the MLLM is a set of latent representations learned from a pre-trained scene encoder that is \textit{jointly trained with the MLLM}, allowing it to more learn grounded feature representations


\section{\ours Framework}
\label{sec:method}

We present \ours, a framework for end-to-end autonomous driving. Given a sequence of multi-view images, the overall objective is to predict the future trajectory of the ego vehicle and answer questions about the scene.   
An overview of \ours is shown in Figure~\ref{fig:main}. The framework has two main components: 1) the vision-based planner;  2) a multi-modal large language model consisting of adapter layers, an LLM, and a series of task-specific decoder heads.

\subsection{Vision-based Planner}
Vision-based end-to-end planners are trained in a multi-task fashion to perform perception, mapping, motion prediction and planning \cite{jiang2023vad,hu2023planning,weng2024drive,tong2023scene}. Considering standard architecture design from \cite{hu2023planning,jiang2023vad,weng2024drive} we decompose the planner into the scene encoder and the planning transformer. The scene encoder provides structured latent representations to the planning transformer that performs waypoint prediction. We call these representations as scene token embeddings. In our framework, we leverage the vision-based planner in two ways. First, it acts as the primary network through which planning is performed, allowing for fast prediction at test time. Second, the scene encoder is shared with the the MLLM, acting as tokenizer to provide it with a highly structured set of inputs. An important distinction of our framework from previous works \cite{hwang2024emma,tian2024tokenize} is that the scene encoder is trained jointly with the MLLM.

\noindent\textbf{Scene encoder}
The scene encoder models the scene as high-dimensional token embeddings that explicitly represent components such as the environment map, the ego vehicle, and the surrounding agents. This is achieved by introducing task specific, learnable query features that are supervised by a series of planning constraints \cite{jiang2023vad}. 
First, a visual backbone encodes the input multi-view image sequence. The obtained feature map is projected into the bird's-eye-view (BEV) space using a set of queries to obtain BEV token embeddings ($B$) \cite{li2022bevformer}. The network learns structured representations of the map and agents in the form of map token embeddings ($M$) and agent token ($A$) embeddings by cross-attending the BEV features with map and agent queries. A randomly initialized learnable embedding is trained to learn the interaction of the ego vehicle with the agent and map components which is the ego query ($E$). We denote the set of these latent scene representations as $BEAM$ token embeddings. These token embeddings are sent to a planning head that predicts future trajectories of the ego and agent vehicles. They are also used as input to the MLLM as multi-modal inputs, as discussed next. 

\subsection{Multi-modal LLM}
\label{sec:MLLM}

 Our objective is to distill knowledge from an MLLM  to the vision-based planner. Towards this, we jointly train an MLLM with the vision-based planner. Specifically, the scene encoder is used as a trainable tokenizer for the MLLM to generate the $BEAM$ token embeddings, as well as used as an encoder for the planning transformer. The advantage from this strategy is two-fold. First, the MLLM receives a highly structured input that captures rich spatio-temporal information relevant to the autonomous driving task. Second, the scene encoder learns features that are grounded in language, improving the robustness of vision-based planning. The main components of the MLLM are adapter layers, the large language model, and a series of task-specific decoder heads which we discuss next.

\subsubsection{Adaptation of Scene Tokens}
In order to project the $BEAM$ token embeddings efficiently while maintaining their distinctiveness, we leverage the query-transformer (Q-former) module \cite{hu2024matryoshka} to compress the visual tokens before being input to the LLM.
We implement component-specific Q-former layers, namely a Map Q-former, a Bird's-Eye-View (BEV) Q-former, an Ego Q-former, and an Agent Q-former. 
Each of these adapters transforms the  representing different modalities into a sequence of fixed-length, higher-dimensional tokens,preparing them for processing by the MLLM.

\subsubsection{MLLM Supervision}

 We design tasks for the MLLM with the objectives of, a) enriching the intermediate scene representations; b) grounding the scene token embeddings in language; and c) training the LLM for planning-related reasoning. Specifically, the MLLM is trained for visual question answering, trajectory estimation, feature distillation, and a set of surrogate tasks \cite{BhattacharyyaP024}.

\myparagraph{Visual question answering. }
The MLLM is trained for visual question answering (VQA) on question-answer pairs of 4 types: perception of the scene, prediction of the agent behavior, prediction of ego-vehicle behavior, and future planning steps \cite{sima2023drivelm}. The answer is predicted based on a multi-modal prompt consisting of the question embeddings and the projected $BEAM$ token embeddings. This is constructed as a standard next-token prediction task and the VQA branch is supervised with a cross-entropy loss as in \cite{li2023blip}. We denote the loss from this task head as $\mathcal{L}_{LLM}$.

\begin{figure}[ht!]
    \centering
    \includegraphics[width=0.8\linewidth]{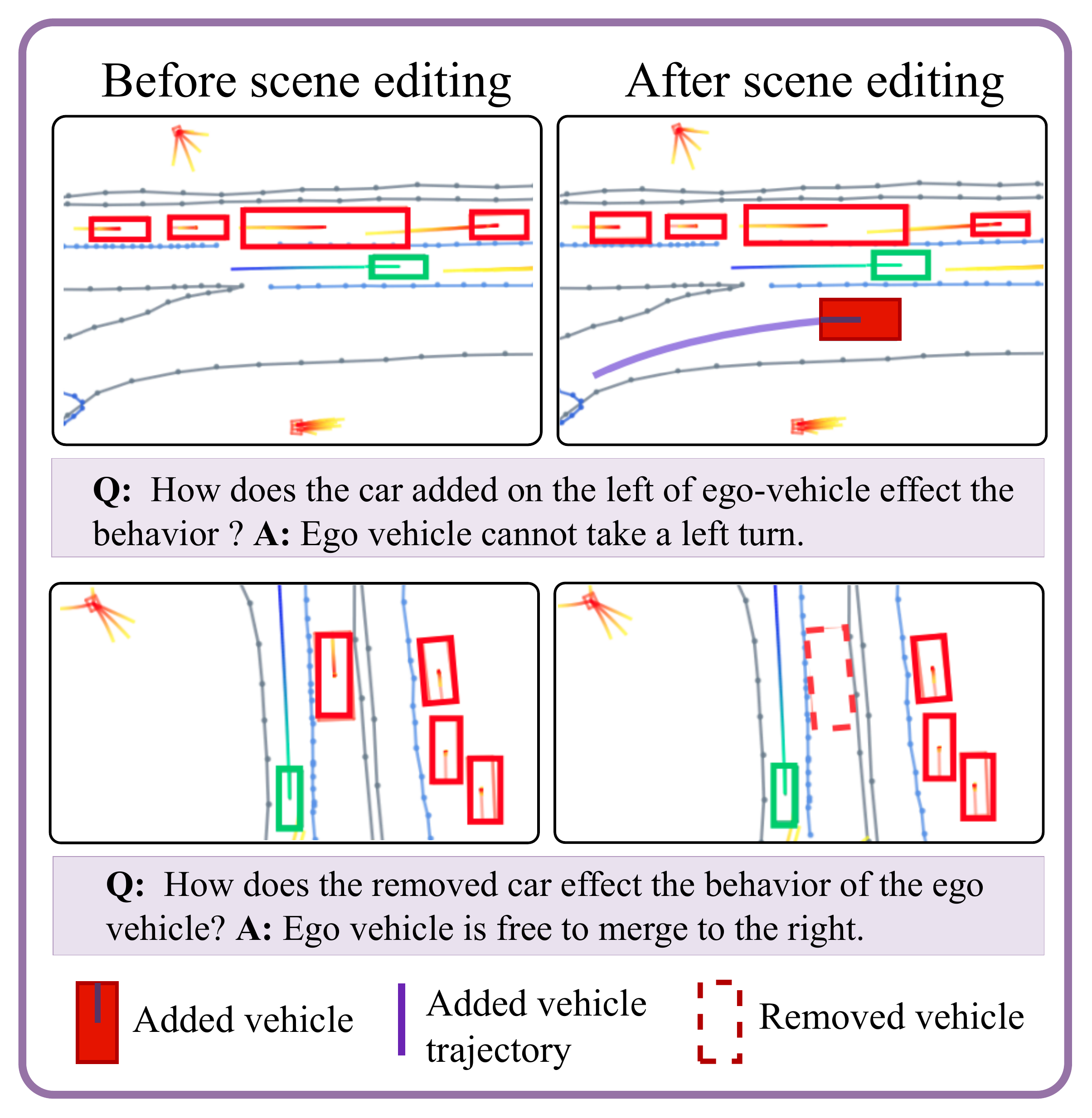}
    \vspace{-5pt}
    \caption{Examples of addition and deletion in scene editing.  In the top row, a car (solid red box) is added on the left of the ego-vehicle (green box). In the bottom row, a car (dashed red box) is removed from the right of the ego-vehicle. A corresponding question-answer pair is created to characterize the edit. }
    \label{fig:scene_editing}
    \vspace{-10pt}
\end{figure}

\myparagraph{Surrogate tasks: }
The main goal of joint training is to enrich the $BEAM$ scene representations learned by the scene encoder. We design MLLM surrogate tasks whose objectives are in line with that of future trajectory prediction.  

\noindent\textit{1) \underline{Masked token reconstruction}:} Each type of token embedding contributes to a holistic representation of the scene. In order to enrich the visual representations, we ask the network to reconstruct a masked BEV input based on the context present in the rest of the multi-modal sequence. We perform random masking after scene encoding and pass the token embeddings to the MLLM. A reconstruction head takes the latent representations from the penultimate layer of LLM and predicts reconstructed BEV token embeddings $\hat{B}$.  This decoder head is supervised with the L2 loss between the prediction and the complete input,
\begin{equation}
    \mathcal{L}_{recon} = ||\hat{B}-B||^2
\end{equation}
where ($\{m(B),E,A,M\}$) is input to the MLLM, and the latent MLLM representations associated with the BEV token embeddings are input to the masked reconstruction decoder head. Here, $m(.)$ denotes random masking.

\noindent\textit{2) \underline{Future BEV prediction}:} An important aspect of planning is anticipating future events. We introduce the surrogate task of future BEV prediction to encourage the LLM to learn spatio-temporal cues useful for planning. Given latent BEV token embeddings, we train a prediction head to predict future BEV token embeddings and supervise this task as the L2 loss between predicted and ground truth future token embeddings:
\begin{equation}
    \mathcal{L}_{future} = ||\hat{F}_t-B_{t+1}||^2 + ||\tilde{F}_t-B_{t+2}||^2
\end{equation}
where $\hat{F}_t$, $\tilde{F}_t$ are the predicted future BEV token embedding at time $t$.  Note, we predict future BEV token embeddings ($B_{t+1}$, $B_{t+2}$) of the multi-view image sequence for time steps $\{t+1, t+2\}$.

\noindent\textit{3) \underline{Scene editing}:} For prediction and reasoning about the ego vehicle, it is crucial to learn how surrounding agents impact the ego-vehicle's future path. We propose a novel scene editing task in which we augment scenes by removing or adding new agents. Along with this, we construct a question-answer pair related to the edit. We show examples of this in Figure~\ref{fig:scene_editing}. For scene addition, given the map constraints, predicted map, the ego bounding box location and the trajectories of predicted agents, we create a way-point trajectory for a new agent of category ``car" or ``truck". 
A new agent token embedding is then created using a linear layer. This new agent token embedding, the corresponding text prompt, and the rest of the $BEAM$ token embeddings are passed as input to the LLM. 
The hidden latent LLM features corresponding to agent token embeddings are then fed into a dedicated scene editing decoder head that performs waypoint prediction of the ego vehicle. The language prediction head performs question-answering on the new QA pair. This task thus contributes to the existing planning constraint loss and VQA loss of the MLLM.

\myparagraph{Distillation} In order to further align the representations learned by the vision planner and the MLLM, we perform knowledge transfer between the penultimate layers of the language model and the planning transformer.
Concretely, we minimize the KL-divergence between the distributions of the hidden features of the planning transformer and the LLM hidden features of the ego-token embeddings:
  \begin{equation}
    \mathcal{L}_{distill} = D_{KL}(P_{llm}||P_{vis})
\end{equation}
where the $P_{vis}$ are the features of penultimate layers of vision planning transformer and $P_{llm}$ are the hidden LLM embeddings features corresponding to the ego-token embeddings of penultimate layers of  MLLM model.  

\noindent\textbf{Loss functions.} 
We supervise the network with a weighted sum of the losses corresponding to planning, visual question answering, distillation, and each surrogate task. 
\begin{equation}
\mathcal{L} =  \mathcal{L} _{planning} + \mathcal{L}_{LLM} + \mathcal{L}_{recon} + \mathcal{L}_{future} + \mathcal{L}_{distill}
\end{equation}
here $\mathcal{L} _{planning}$ comes from the planning objective used to train \cite{hu2023planning,jiang2023vad}. Here, the loss weights are chosen to bring each value to the same scale.

\begin{table*}[ht!]
    \centering
    \caption{Comparison of L2 trajectory error and collision rate on nuScenes \cite{caesar2020nuscenes} using standardized evaluation \cite{weng2024drive}. Models are evaluated on the general validation split as well as a ``targeted'' split of challenging samples from \cite{weng2024drive}. The performance of the \ours~model variants are in shades of purple. We summarize results by averaging over at $t=\{1,2,3\}s$ as well as at all time steps. The best average performance in each setting is in bold. ``+'' indicates the use of ego-status information. }
    \vspace{-5pt}
    \label{tab:PARADrive_eval_test_nuscenes}
    \resizebox{1\linewidth}{!}{
    \begin{tabular}{lccccccccccc}
    \hline
     \multicolumn{1}{c}{Method} & Using & \multicolumn{5}{c}{Traj L2 (m) $\downarrow$} & \multicolumn{5}{c}{Collision (\%) $\downarrow$} \\ \cline{3-12}
      & Ego status & 1s & 2s & 3s & \cellcolor{red!25}Ave$_{1,2,3s}$ & \cellcolor{red!25}Ave$_{all}$  & 1s & 2s & 3s & \cellcolor{red!25}Ave$_{1,2,3s}$ & \cellcolor{red!25}Ave$_{all}$ \\ \hline
    \rowcolor{gray!20} \multicolumn{12}{c}{\textbf{Full validation split}} \\ 
    UniAD\cite{hu2023planning} & \bccross & 0.48 & 0.89 & 1.47 & 0.95 & 0.83 & 0.32 & 0.29 & 0.73 & 0.45 & 0.40 \\
    VAD-Base \cite{jiang2023vad} & \bccross & 0.41  & 0.86 & 1.46 & 0.91 & 0.78 & 0.02 & 0.26 & 0.83 & 0.37 & 0.30 \\
    PARA-Drive\cite{weng2024drive} & \bccross & 0.26 & 0.59 & 1.12 & 0.66 & 0.56 & 0.00 & 0.12 & 0.65 & 0.26 & 0.17 \\
    TOKEN\cite{tian2024tokenize} & \bccross & 0.26 & 0.70 & 1.46 & 0.81 & 0.68 & -- & -- & -- & -- & 0.15 \\
    \ccb \ours (VAD-Tiny) &\ccb\bccross &\ccb 0.20 &\ccb 0.53 &\ccb 1.10 &\ccb 0.61 &\ccb 0.51 &\ccb 0.00 &\ccb 0.07 &\ccb 0.19 &\ccb 0.09 &\ccb \textbf{0.06} \\
    \ccdb \ours (VAD-Base) &\ccdb \bccross &\ccdb 0.18 &\ccdb 0.50 &\ccdb 1.02 &\ccdb \textbf{0.57} &\ccdb \textbf{0.47} &\ccdb 0.00 &\ccdb 0.05 &\ccdb 0.16 &\ccdb \textbf{0.07} &\ccdb \textbf{0.06} \\
     AD-MLP\cite{zhai2023rethinking} & \bccheck & 0.23 & 0.58 & 1.18 & 0.66 & 0.56 & 0.00 & 0.14 & 0.70 & 0.28 & 0.20 \\
    PARA-Drive+\cite{weng2024drive} & \bccheck & 0.20 & 0.52 & 1.04 & 0.59 & 0.49 & 0.00 & 0.09 & 0.49 & 0.19 & 0.13 \\
    \ccb \oursp (VAD-Tiny) &\ccb \bccheck &\ccb 0.19 &\ccb 0.51 &\ccb 1.06 &\ccb 0.59 &\ccb 0.49 &\ccb 0.00 &\ccb 0.06 &\ccb 0.16 &\ccb 0.08 &\ccb \textbf{0.06} \\
    \ccdb \oursp (VAD-Base) &\ccdb \bccheck &\ccdb 0.18 &\ccdb 0.48 &\ccdb 1.01 &\ccdb \textbf{0.56} &\ccdb \textbf{0.46} &\ccdb 0.00 &\ccdb 0.05 &\ccdb 0.16 &\ccdb \textbf{0.07} &\ccdb \textbf{0.06} \\
    \ccnb \ours-Dual+ (VAD-Tiny) &\ccnb \bccheck &\ccnb 0.18 &\ccnb 0.50 &\ccnb 1.03 &\ccnb 0.57 &\ccnb 0.47 &\ccnb 0.00 &\ccnb 0.05 &\ccnb 0.16 &\ccnb 0.08 &\ccnb \textbf{0.06} \\ \hline
    
    \rowcolor{gray!20} \multicolumn{12}{c}{\textbf{Targeted validation split}} \\
    UniAD\cite{hu2023planning} & \bccross & 0.47 & 1.09 & 1.92 & 1.16 & 0.99 & 0.00 & 0.00 & 0.73 & 0.24 & 0.15 \\
    VAD-Base \cite{jiang2023vad} & \bccross & 0.53 & 1.20 & 2.07 & 1.27 & 1.08 & 0.00 & 0.29 & 0.87 & 0.39 & 0.34 \\
    PARA-Drive\cite{weng2024drive} & \bccross & 0.38 & 0.97 & 1.88 & 1.08 & 0.91 & 0.00 & 0.00 & 0.72 & 0.24 & 0.14 \\
     \ccb \ours (VAD-Tiny) &\ccb  \bccross&\ccb 0.32 &\ccb 0.88 &\ccb 1.70 &\ccb 0.97 &\ccb 0.81 &\ccb 0.00 &\ccb 0.03 &\ccb 0.24 &\ccb 0.09 &\ccb \textbf{0.05} \\
    \ccdb \ours (VAD-Base) & \ccdb \bccross &\ccdb 0.29 &\ccdb 0.79 &\ccdb 1.49 &\ccdb \textbf{0.83} &\ccdb \textbf{0.71} &\ccdb 0.00 &\ccdb 0.03 &\ccdb 0.22 &\ccdb \textbf{0.08} &\ccdb \textbf{0.05} \\
    AD-MLP\cite{zhai2023rethinking} & \bccheck & 0.33 & 0.99 & 2.06 & 1.13 & 0.94 & 0.00 & 0.58 & 3.62 & 1.40 & 0.94 \\
    PARA-Drive+ \cite{weng2024drive}& \bccheck & 0.29 & 0.76 & 1.46 & 0.83 & 0.70 & 0.00 & 0.00 & 0.29 & 0.10 & 0.05 \\
    \ccb \oursp (VAD-Tiny) &\ccb \bccheck &\ccb 0.31 &\ccb 0.83 &\ccb 1.67 &\ccb 0.94 &\ccb 0.77  &\ccb 0.00 &\ccb 0.04 &\ccb 0.17 &\ccb 0.07 &\ccb 0.05 \\
    \ccdb \oursp (VAD-Base) &\ccdb \bccheck &\ccdb 0.27 &\ccdb 0.74 &\ccdb 1.38 &\ccdb \textbf{0.79} &\ccdb \textbf{0.64} &\ccdb 0.00 &\ccdb 0.03 &\ccdb 0.16 &\ccdb \textbf{0.06} &\ccdb \textbf{0.04} \\
    \ccnb \ours-Dual+ (VAD-Tiny) &\ccnb \bccheck &\ccnb 0.28 &\ccnb 0.76 &\ccnb 1.49 &\ccnb 0.84 &\ccnb 0.69 &\ccnb 0.00 &\ccnb 0.03 &\ccnb 0.17 &\ccnb 0.07 &\ccnb 0.05 \\
    
    \hline
    \end{tabular}
    }
    \vspace{-10pt}
\end{table*}

\section{Experimental Setup}
\label{sec:experiments}
 In Section \ref{sec:arch}, we present the architecture design details, in Section \ref{sec:train} we present the training strategy, and in Section \ref{sec:data}, we provide information about the datasets used. Importantly, in Section \ref{sec:eval_strat} we detail multiple evaluation strategies for fair and comprehensive comparison with existing works.

\subsection{Architecture design}
\label{sec:arch}
\noindent\textbf{Vision-based planner}
We conduct experiments using two vision-based end-to-end planners, VAD \cite{jiang2023vad} and UniAD \cite{hu2023planning}  due to their performance, efficiency, and ability to model the interaction between scene components. We use two model size variants of VAD. Both planners are trained for perception, motion prediction and planning, while UniAD also performs occupancy prediction.

\noindent\textbf{MLLM design}
The MLLM is made up of adapter layers, a language model, and a set of task-specific decoder layers.
We project each $BEAM$ token embedding with its own dedicated Q-former adapter layer following \cite{hu2024matryoshka}. After projection, the $BEAM$ token embeddings and language token embeddings lie in the same embedding space. In order to reduce memory consumption, we limit the agent token sequence length to a fraction of the total number of agents and perform sequence-wise upsampling after input to the language model. The adapter layers are randomly initialized. 
We use the LLM from LLaVA-v1.5-7B \cite{liu2024visual} as our language model base. The ego prediction and agent motion prediction task heads are multi-layer-perceptron networks following \cite{jiang2023vad}. For the surrogate task decoder heads, we use 3 Linear layers with a ReLU activation layer.

\subsection{Training}
\label{sec:train}
\ours~training follows a two-stage approach. First the vision-only planner is pre-trained for perception, prediction, and planning for 60 epochs in order to learn informative latent scene representations. Second, we perform joint training of the vision planner and the MLLM for an additional 30 epochs, incorporating all proposed tasks and losses detailed in Section \ref{sec:method}. 
In the second stage, the language model of the MLLM is fine-tuned using LoRA \cite{hu2021lora}. 

Please refer to Section \ref{sec:training} of the appendix for more details. 

\subsection{Datasets}
\label{sec:data}
We use the nuScenes dataset for the task of open-loop planning \cite{caesar2020nuscenes}. This dataset consists of 28k total samples in a 22k/6k training/validation split. The objects in each scene are annotated with 3D bounding box, orientation, and vehicle speed information. Additionally, we make use of the CAN bus information for the ego vehicle's  ground-truth trajectory annotations. For supervising visual question answering, we train with DriveLM \cite{sima2023drivelm}. This dataset consists of a 4k subset of samples from the nuScenes dataset annotated with 300k QA pairs related to perception, prediction, planning, and behavior of the ego vehicle. For the samples in nuScenes that do not have text annotations, we create a set of perception, planning, and prediction QA pairs based on the numerical annotations. 

We use this information to prompt a Llama3-70B model \cite{dubey2024llama} with in-context examples to generate DriveLM-like QA pairs. For the behavior QA pair, we create a rule-based algorithm to categorize the future movement of the ego-vehicle based on the ground-truth trajectory values and ego vehicle speed. More details on this method as well as examples of generated QA pairs may be found in Section \ref{sec:data_gen} of the appendix.

\subsection{Evaluation details}
\label{sec:eval_strat}
We evaluate planning performance based on the predicted future trajectories of the ego vehicle 3 seconds into the future, where 2 waypoints are predicted per second. We use two metrics, the L2 error (in meters) between the predicted and ground-truth waypoints as well as the collision rate (in \%) between the ego-vehicle and the surrounding vehicles. 

\noindent\textbf{Standardized vs VAD evaluation.}  In \cite{weng2024drive}, Weng \etal point out inconsistencies in planning evaluation across VAD \cite{jiang2023vad}, UniAD \cite{hu2023planning} and AD-MLP \cite{zhai2023rethinking}, namely in the way L2 error is averaged over time and the way noisy and invalid frames are considered. The authors also point out that collision performance is considerably improved by using a more finely discretized BEV map. Accounting for these inconsistencies, they propose a ``standardized'' evaluation metric to maintain a fair comparison across methods. For accurate reporting and fair comparison with \cite{weng2024drive,tian2024tokenize}, we use this standardized metric to evaluate our model. 
Certain existing works use the evaluation scheme followed by \cite{jiang2023vad} (which we call VAD evaluation) and do not make code or models available \cite{tian2024drivevlm,wang2024omnidrive,pan2024vlp,hu2022st}. Here, we compare reported results against \ours~evaluated using VAD evaluation.

\begin{table*}[t!]
\centering
\caption{Comparison of L2 trajectory error and collision rate on nuScenes \cite{caesar2020nuscenes} using VAD evaluation \cite{jiang2023vad}. Models are evaluated on the general validation split. The performance of the \ours~model variants are in shades of purple. The performance of the MLLM-branch \ours~model variant is in blue. We summarize results by averaging over all time steps. The best average performance in each setting is in bold. ``+'' indicates the use of ego-status information.}
\vspace{-5pt}
\label{tab:VAD_eval_test_nuscenes}
\resizebox{1\linewidth}{!}{
\begin{tabular}{lccccccccccc}
\hline
\multicolumn{1}{c}{Method} & Using & \multicolumn{4}{c}{Traj L2 (m) $\downarrow$} & \multicolumn{4}{c}{Collision (\%) $\downarrow$} & Latency (ms) & FPS \\  \cline{3-12}
& Ego status & 1s & 2s & 3s & \cellcolor{red!25}Ave$_{all}$ & 1s & 2s & 3s & \cellcolor{red!25}Ave$_{all}$ &  &  \\ \hline
ST-P3 \cite{hu2022st} & \bccross & 1.33 & 2.11 & 2.90 & 2.11 & 0.23 & 0.62 & 1.27 & 0.71 & 628.3 & 1.6 \\
UniAD\cite{hu2023planning} & \bccross & 0.48 & 0.96 & 1.65 & 1.03 & 0.05 & 0.17 & 0.71 & 0.31 & 555.6 & 1.8 \\
VAD-Tiny\cite{jiang2023vad} & \bccross & 0.46 & 0.76 & 1.12 & 0.78 & 0.21 & 0.35 & 0.58 & 0.38 & 59.5 & 16.8 \\
VAD-Base\cite{jiang2023vad} & \bccross & 0.41 & 0.70 & 1.05 & 0.72 & 0.07 & 0.17 & 0.41 & 0.22 & 224.3 & 4.5 \\ 
VLP-VAD-Base\cite{pan2024vlp} & \bccross & 0.26 & 0.47 & 0.78 & 0.50 & 0.12 & 0.17 & 0.42 & 0.23 & -- & -- \\ 
OmniDrive \cite{wang2024omnidrive} & \bccross & 0.40 & 0.80 & 1.32 & 0.84 & 0.04 & 0.46 & 2.32 & 0.94 & -- & -- \\ 
\ccb \ours(UniAD) &\ccb \bccross &\ccb 0.15 &\ccb 0.30 &\ccb 0.56 &\ccb 0.34 &\ccb 0.06 &\ccb 0.08 &\ccb 0.22 &\ccb 0.12 &\ccb 560 &\ccb 1.8 \\
\ccdb \ours(VAD-Tiny) &\ccdb \bccross &\ccdb 0.18 &\ccdb 0.36 &\ccdb 0.61 &\ccdb 0.38 &\ccdb 0.07 &\ccdb 0.10 &\ccdb 0.27 &\ccdb 0.15 &\ccdb 59.5 &\ccdb 16.8 \\
\ccnb \ours(VAD-Base) &\ccnb \bccross&\ccnb 0.13 &\ccnb 0.27 &\ccnb 0.47 &\ccnb \textbf{0.29} &\ccnb 0.05 &\ccnb 0.08 &\ccnb 0.16 &\ccnb \textbf{0.10} &\ccnb 226 &\ccnb 4.5 \\ \hline
VAD-Tiny & \bccheck & 0.20 & 0.38 & 0.65 & 0.41 & 0.10 & 0.12 & 0.27 & 0.16 & 59.5 & 16.8 \\
VAD-Base & \bccheck & 0.17 & 0.34 & 0.60 & 0.37 & 0.07 & 0.10 & 0.24 & 0.14 & 224.3 & 4.5 \\
DriveVLM \cite{tian2024drivevlm}& \bccheck & 0.18 & 0.34 & 0.68 & 0.40 & 0.10 & 0.22 & 0.45 & 0.27 & -- & -- \\
DriveVLM-Dual (VAD-Base) \cite{tian2024drivevlm}& \bccheck & 0.15 & 0.29 & 0.48 & 0.31 & 0.05 & 0.08 & 0.17 & 0.10 & -- & -- \\

\ccdb \ours+~(VAD-Tiny) &\ccdb \bccheck &\ccdb 0.17	 &\ccdb  0.33	&\ccdb 0.59 &\ccdb	0.37	  &\ccdb 0.07	 &\ccdb 0.10	 &\ccdb 0.26	 &\ccdb 0.14&\ccdb 59.5 &\ccdb 16.8 \\
\ccnb \ours+~(VAD-Base) &\ccnb \bccheck &\ccnb 0.12 &\ccnb 0.25 &\ccnb 0.44 &\ccnb \textbf{0.27} &\ccnb 0.04 &\ccnb 0.06 &\ccnb 0.15 &\ccnb \textbf{0.08} &\ccnb 226 &\ccnb 4.5  \\ 
\cclc \ours-Dual+~(VAD-Tiny) &\cclc \bccheck &\cclc 0.14 &\cclc 0.27 &\cclc 0.46 &\cclc 0.29 &\cclc 0.05 &\cclc 0.07 &\cclc 0.15 &\cclc 0.09 &\cclc 286 &\cclc 3.5  \\\hline
\end{tabular}}
\end{table*}

\begin{table}[ht!]
\caption{Long-tail performance comparison of L2 trajectory error and collision rate on nuScenes \cite{caesar2020nuscenes} validation set using standardized evaluation \cite{weng2024drive}. The performance of vision planner \ours~model variants are in shades of purple. The performance of MLLM-branch \ours~model variants are in shades of blue. We summarize results by averaging over at $t=\{1,2,3\}s$ as well as at all time steps. The best average performance in each setting is in bold. }
\vspace{-5pt}
\label{tab:long_tail}
\resizebox{.985\linewidth}{!}{
\begin{tabular}{lcccccc}
\hline
\multicolumn{1}{c}{\textbf{Method}} & \multicolumn{5}{c}{Traj L2 (m)$\downarrow$} & \multicolumn{1}{c}{Coll. (\%) $\downarrow$} \\ \cline{2-7} 
 & 1s & 2s & 3s & \cellcolor{red!25}Ave$_{1,2,3s}$ & \cellcolor{red!25}Ave$_{all}$ & \cellcolor{red!25}Ave$_{all}$ \\ \hline

\rowcolor{gray!20} \multicolumn{7}{c}{\textbf{3-point turn (zero-shot)}} \\
VAD-Tiny & 0.76 & 1.68 & 3.04 & 1.83 & 1.55 & 0.00 \\
VAD-Base & 0.71 & 1.66 & 3.24 & 1.87 & 1.57 & 0.00 \\
PARA-Drive \cite{weng2024drive} & 0.50 & 1.38 & 2.76 & 1.55 & 1.29 & 5.33 \\
TOKEN \cite{tian2024tokenize} & 0.39 & 1.29 & 2.60 & 1.43 & 1.18 & 4.00 \\
\ccb \ours~(VAD-Tiny) & \ccb 0.47 & \ccb 1.27 & \ccb 2.55 & \ccb 1.43 & \ccb 1.17 & \ccb 0.00 \\
\ccdb \ours~(VAD-Base) & \ccdb 0.36 & \ccdb 1.18 & \ccdb 2.37 & \ccdb 1.30 & \ccdb 1.05 & \ccdb 0.00 \\
\cclc \ours~ (MLLM) & \cclc 0.38 & \cclc 1.12 & \cclc 2.35 & \cclc 1.28 & \cclc \textbf{1.04} & \cclc 0.00 \\
\ccdc \ours-Dual (VAD-Tiny) & \ccdc 0.38 & \ccdc 1.08 & \ccdc 2.35 & \ccdc \textbf{1.27} & \ccdc \textbf{1.04} & \ccdc 0.00 \\ 
\hline

\rowcolor{gray!20} \multicolumn{7}{c}{\textbf{Resume from stop}} \\
VAD-Tiny & 0.64 & 1.63 & 2.99 & 1.75 & 1.49 & 0.00 \\
VAD-Base & 0.60 & 1.72 &  2.83 & 1.72 & 1.42 & 0.00\\
PARA-Drive & 0.14 & 0.79 & 2.30 & 1.08 & 0.85 & 0.00 \\
TOKEN & 0.13 & 0.70 & 1.58 & 0.80 & \textbf{0.65} & 0.00 \\
\ccb \ours~(VAD-Tiny) & \ccb 0.26 & \ccb 1.02 & \ccb 2.40 & \ccb 1.23 & \ccb 0.99 & \ccb 0.00 \\
\ccdb \ours (VAD-Base) & \ccdb 0.15 & \ccdb 0.65 & \ccdb 1.34 & \ccdb \textbf{0.71} & \ccdb 0.66 & \ccdb 0.00 \\
\cclc \ours~ (MLLM) & \cclc 0.24 & \cclc 1.02 & \cclc 2.16 & \cclc 1.14 & \cclc 0.93 & \cclc 0.00 \\ 
\ccdc \ours-Dual (VAD-Tiny) & \ccdc 0.24 & \ccdc 0.98 & \ccdc 2.13 & \ccdc 1.11 & \ccdc 0.91 & \ccdc 0.00 \\ 
\hline

\rowcolor{gray!20} \multicolumn{7}{c}{\textbf{Overtake}} \\
VAD-Tiny & 0.58 & 1.27 & 2.12 & 1.32 & 1.14 & 2.42 \\
VAD-Base & 0.46 & 1.16 & 2.17 & 1.26 & 1.06 & 2.49 \\
PARA-Drive & 0.27 & 0.89 & 1.94 & 1.03 & 0.85 & 2.30 \\
TOKEN & 0.29 & 0.77 & 1.63 & 0.90 & 0.74 & 0.00 \\
\ccb \ours~(VAD-Tiny) & \ccb 0.24 & \ccb 0.75 & \ccb 1.49 & \ccb 0.83 & \ccb 0.69 & \ccb 1.32 \\
\ccdb \ours~(VAD-Base) & \ccdb 0.24 & \ccdb 0.72 & \ccdb 1.50 & \ccdb \textbf{0.82} & \ccdb \textbf{0.66} & \ccdb \textbf{1.29} \\
\cclc \ours~ (MLLM) & \cclc 0.24 & \cclc 0.73 & \cclc 1.50 & \cclc \textbf{0.82} & \cclc 0.67 & \cclc 1.30 \\ 
\ccdc \ours-Dual (VAD-Tiny) & \ccdc 0.24 & \ccdc 0.73 & \ccdc 1.50 & \ccdc \textbf{0.82} & \ccdc 0.67 & \ccdc 1.30 \\ 
\hline
\end{tabular}}
\vspace{-15pt}
\end{table}

\noindent\textbf{Validation splits.} We evaluate planning performance across 3 types of splits from the nuScenes validation set: 1) the entire validation set of 6019 samples 2) a 689 sample sub-set more difficult ``targeted'' scenarios in which the ego-vehicle turns during navigation \cite{weng2024drive}, and 3) long-tail events in which the ego-vehicle encounters rare maneuvers \cite{tian2024tokenize}.

\section{Experimental Results}
\label{sec:results}

We perform a comprehensive evaluation of \ours~ on the nuScenes open-loop planning benchmark in a variety of evaluation settings. We compare our performance against that of recent vision-based planners \cite{hu2022st,hu2023planning,jiang2023vad,weng2024drive} as well as recent MLLM planners \cite{pan2024vlp,wang2024omnidrive,tian2024drivevlm,tian2024tokenize}.
To facilitate a fair comparison of \ours~ with recent works, we use two evaluation strategies VAD evaluation~\cite{jiang2023vad} and standardized evaluation~\cite{weng2024drive}. Please refer to Section \ref{sec:eval_strat} for more details.

\noindent\textbf{Standardized evaluation.} Under the standardized evaluation strategy, we evaluate each method on two validation splits: 1) the full nuScenes validation split, and 2) a ``targeted'' subset of the nuScenes validation split that contains scenarios where the ego-vehicle must take a turn, as established in \cite{weng2024drive}.  We present our results in Table \ref{tab:PARADrive_eval_test_nuscenes}. 
The baseline vision planners VAD \cite{jiang2023vad} and UniAD \cite{hu2023planning} exhibit sub-optimal performance in more difficult targeted scenarios. \ours~demonstrates significant improvements over baseline methods, achieving a $35\% $ reduction in L2 trajectory error and an $80\%$ reduction in collision rate on the overall nuScenes validation set, and  
a $35\%$ reduction in L2 trajectory error in the ``targeted'' split. More notably, we outperform recent state-of-the-art methods PARA-Drive \cite{weng2024drive} and TOKEN \cite{tian2024tokenize} consistently across both validation splits. We observe a strong improvement in collision rate, improving on that of TOKEN by $60\%$. Unlike TOKEN, our approach does not rely on an LLM for inference, thus making \ours~more efficient as well as more accurate.

\noindent\textbf{VAD evaluation.}
For a fair comparison with recent works that use the evaluation strategy from VAD \cite{jiang2023vad}, we present an additional set of results under this strategy. \footnote{We compare with reported results due to lack of published code.}  In Table~\ref{tab:VAD_eval_test_nuscenes}, we present the evaluation of \ours~on the nuScenes validation set \cite{caesar2020nuscenes}. We significantly outperform the baseline vision-based planners VAD \cite{jiang2023vad} and UniAD \cite{hu2023planning}. Notably, \ours~built on VAD-Tiny outperforms VAD-Base by $47\%$ in L2 trajectory error while being 4 times faster. We also outperform MLLM planners like OmniDrive \cite{wang2024omnidrive} and DriveVLM \cite{tian2024drivevlm} without requiring an LLM at inference, making our approach more accurate as well as efficient.

\begin{figure*}[t!]
    \centering
    \begin{subfigure}{0.49\textwidth}
        \includegraphics[width=\linewidth]{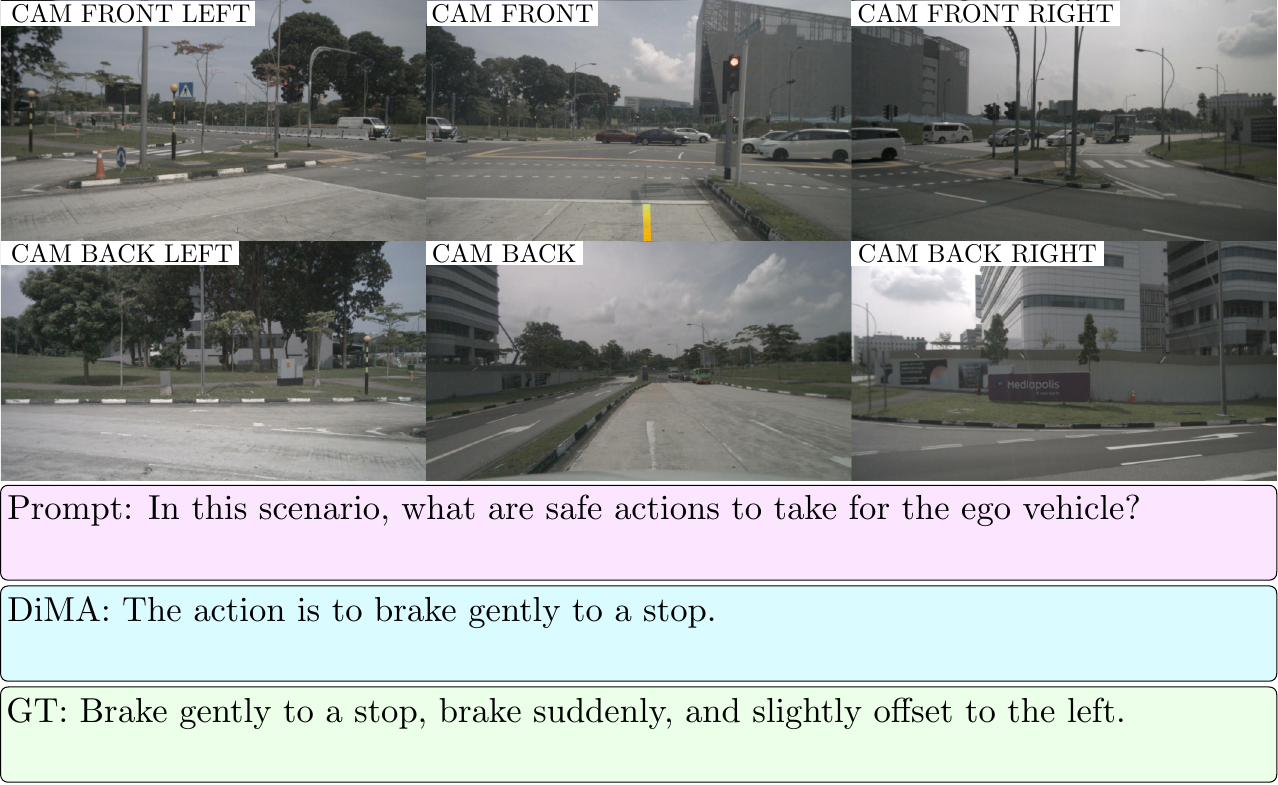}
    \end{subfigure}
    \hfill
    \begin{subfigure}{0.49\textwidth}
        \includegraphics[width=\linewidth]{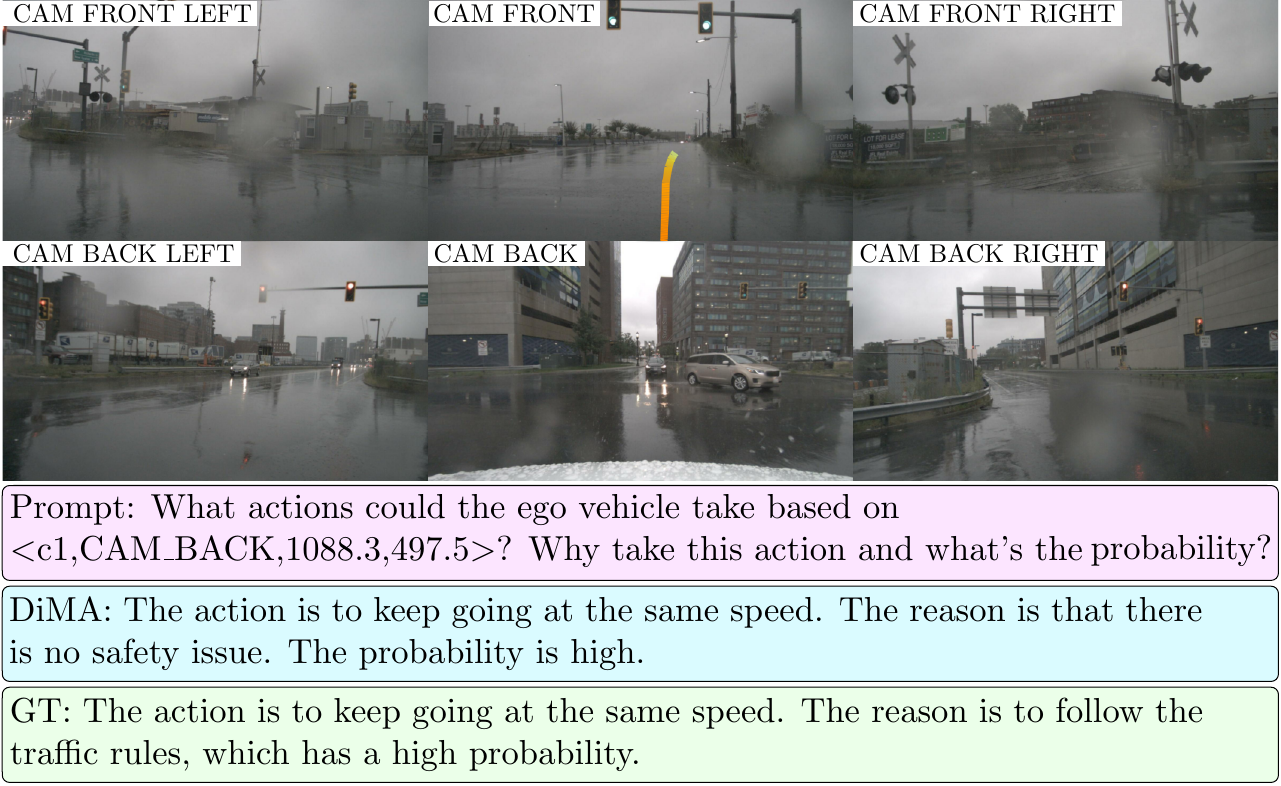}
    \end{subfigure}
    \vspace{-5pt}
    \caption{Visualization of planning performance by the MLLM branch of \ours-VAD-Tiny. We plot the predicted trajectory (orange-yellow) and show an example response of the language model branch to a question from the DriveLM test dataset \cite{sima2023drivelm}.}
    \label{fig:vqa_qual}
    \vspace{-8pt}
\end{figure*}

\begin{table*}[t!]
\caption{Results of ablation experiments of \ours-VAD-Tiny on nuScenes under VAD evaluation. The baseline model is VAD-Tiny. We report L2 trajectory error and collision rate averaged over all time steps.}
\vspace{-5pt}
\label{tab:ablation}
\resizebox{.985\linewidth}{!}{
\begin{tabular}{lccccccccccccccc}
\hline
\multirow{2}{*}{ID} & \multirow{2}{*}{VQA} & \multirow{2}{*}{\begin{tabular}[c]{@{}c@{}}Scene \\ tokens\end{tabular}} & \multirow{2}{*}{Distillation} & \multirow{2}{*}{\begin{tabular}[c]{@{}c@{}}LLM \\ Planning\end{tabular}} & \multicolumn{3}{c}{Surrogate tasks}                                                                                                                                           & \multicolumn{4}{c}{Traj L2 (m) $\downarrow$} & \multicolumn{4}{c}{Collision (\%) $\downarrow$} \\ \cline{6-16} 
                  &   &                                                                          &                               &                                                                          & \begin{tabular}[c]{@{}c@{}}Masked \\ recon.\end{tabular} & \begin{tabular}[c]{@{}c@{}}Future \\ pred.\end{tabular} & \begin{tabular}[c]{@{}c@{}}Scene \\ editing\end{tabular} & 1s      & 2s    & 3s    & \cellcolor{red!25}Ave$_{all}$   & 1s     & 2s      & 3s     & \cellcolor{red!25}Ave$_{all}$    \\ \hline
1 & \ccross                 & \ccross                                                                       & \ccross                            & \ccross                                                                       & \ccross                                                       & \ccross                                                      & \ccross                                                       & 0.33    & 0.58  & 0.89  & 0.60  & 0.12   & 0.19    & 0.55   & 0.29  \\
2 & \ccheck                  & BEV                                                                      &     \ccross                           & \ccheck                                                                      &                         \ccross                                 &                  \ccross                                      &         \ccross                                                 & 0.36    & 0.60  & 0.91  & 0.62  & 0.10    & 0.17    & 0.52   & 0.26   \\
3 & \ccheck                  & BEV, Map                                                                 & \ccross                            & \ccheck                                                                      & \ccross                                                       & \ccross                                                      & \ccross                                                       & 0.28    & 0.52  & 0.87  & 0.56  & 0.10   & 0.17    & 0.49   & 0.25   \\
4 & \ccheck                  & All                                                               & \ccross                            & \ccheck                                                                      & \ccross                                                       & \ccross                                                     & \ccross                                                       & 0.26    & 0.51  & 0.83  & 0.52  & 0.10   & 0.16    & 0.38   & 0.21   \\
5 & \ccheck                  & All                                                               & \ccheck                           & \ccheck                                                                      & \ccross                                                       & \ccross                                                      & \ccross                                                       & 0.22   & 0.44  & 0.77  & 0.48  & 0.09 & 0.14 & 0.34 & 0.19   \\
6 & \ccheck                  & All                                                                & \ccheck                           & \ccheck                                                                      & \ccheck                                                      & \ccross                                                      & \ccross                                                       & 0.19    & 0.39  & 0.68  & 0.42  & 0.09 & 0.13 &  0.32 & 0.18   \\
7 & \ccheck                  & All                                                                & \ccheck                           & \ccheck                                                                      & \ccheck                                                      & \ccheck                                                     & \ccross                                                       & 0.18    & 0.37  & 0.63  & 0.39  &  0.07 & 0.11 & 0.29& 0.16 \\
8 & \ccheck                  & All                                                                & \ccheck                           & \ccheck                                                                      & \ccheck                                                      & \ccheck                                                     & \ccheck                                                      & 0.18    & 0.36  & 0.61  & 0.38  & 0.07 & 0.10 & 0.27 & 0.15 \\ \hline
\end{tabular}
}
\vspace{-5pt}
\end{table*}

\noindent\textbf{Performance in long-tail scenarios.} Autonomous driving performance is particularly crucial for rare scenarios. Using the manually selected long-tail events from \cite{tian2024tokenize}, we demonstrate the robustness of \ours~in cases of novel navigation and perception. In  Table~\ref{tab:long_tail}, we evaluate the performance of \ours~for three long-tail scenarios, ``3-point turn'',  ``resume from the stop'' and ``overtake'', selected from the nuScenes validation set. Training with \ours~results in a significant and consistent boost in performance from the baseline vision-based planners. We also consistently outperform PARA-Drive and TOKEN. Notably, the ``3-point turn'' case is a zero-shot scenario, not present in the training data. We achieve the lowest L2 trajectory error for this case.

\noindent\textbf{Evaluating the MLLM branch.}
We also evaluate the MLLM branch of \ours~ and demonstrate its planning and visual question-answering capabilities. We perform LLM-based planning using \ours~in two ways. First, we directly evaluate the waypoint predictions of the MLLM planning decoder head. This is denoted by \ours~(MLLM). Secondly, we define \ours-Dual, an optional hybrid inference model that combines the strengths of both the vision and MLLM branches. We perform max-pooling on the penultimate layer features of both vision planning transformer and MLLM ego prediction. The resulting pooled features are then fed back into both the vision planning transformer and MLLM ego prediction for waypoint prediction. This max pooling encourages consistency between the predictions of the two models, leading to a more robust and potentially improved overall prediction as seen in Table \ref{tab:long_tail}.  Using the MLLM at inference results in a significant boost in performance, with \ours-Dual (VAD-Tiny) matching the performance of \ours~(VAD-Base) in Table \ref{tab:VAD_eval_test_nuscenes}.

\noindent\textbf{VQA results.}
We provide qualitative examples of visual question-answering in the DriveLM dataset \cite{sima2023drivelm} in Figure \ref{fig:vqa_qual}. We demonstrate the ability of \ours~to reason about safe actions to be taken by the ego vehicle. We provide a more examples in the Section \ref{sec:vqa} of the appendix.

\subsection{Ablation study.}
We examine the role of different design choices and proposed tasks in the \ours~framework through a comprehensive ablation study in Table \ref{tab:ablation}.
 In order to do so, we build our final frame-work step-by-step. We begin with VAD-Tiny as our baseline model in the first row, denoted by ID-1. In ID-2, we observe inconsistent gains and drops in performance by naively training an LLM with the VAD-Tiny under a the planning and VQA objectives while using only BEV features. 
 By comparing ID-3 and ID-4, we show that that providing more scene context to the MLLM by using all the $BEAM$ token embeddings to train both LLM and vision-based planner benefits the performance, resulting a significant improvement in trajectory estimation and collision rate. 
 We show in ID-5 that explicitly distilling the information from the penultimate layers of the MLLM to the planning transformer gives another boost in performance. In ID-6, ID-7 and ID-8 we demonstrate the importance of surrogate tasks like masked-reconstruction, future prediction and scene editing in helping the scene encoder to learn richer representations useful for planning.

\section{Conclusion}
\label{sec:conclusion}

We introduce \ours, an end-to-end autonomous driving framework for robust and efficient planning. \ours~ distills knowledge from an MLLM to a vision-based planner, employing a novel joint training strategy alongside carefully designed surrogate tasks such as masked token reconstruction, future token prediction, and scene editing. \ours~enables the model to learn semantically grounded scene representations, improving its performance in difficult navigation scenarios. We conduct extensive evaluations and ablation studies, showing that training with \ours~ results in a planner that excels in long-tail cases while maintaining computations efficiency.

{
    \small
    \bibliographystyle{ieeenat_fullname}
    \bibliography{main}
}

\onecolumn 
\begin{center}
    \Large{\textbf{Appendix}}
\end{center} 
\begin{appendix}
    
\noindent We present the appendix for the paper ``Distilling Multi-modal Large Language Models for Autonomous Driving''. In Section \ref{sec:method}, we provide details on the surrogate tasks module of \ours~. Section \ref{sec:exp} has the details on the training setup and the procedure for generating additional text annotations for the nuScenes dataset. In Section \ref{sec:qual}, we present additional qualitative results of \ours's vision-based planning branch as well as visual question-answering results from the MLLM branch. We provide an additional quantitative evaluation of \ours~ in Section \ref{sec:quant}.


\section{Surrogate tasks overview}\label{sec:surrogate}

\begin{wrapfigure}{r}{0.4\textwidth}
    \centering
    \vspace{-10pt}
    \includegraphics[width=1\linewidth]{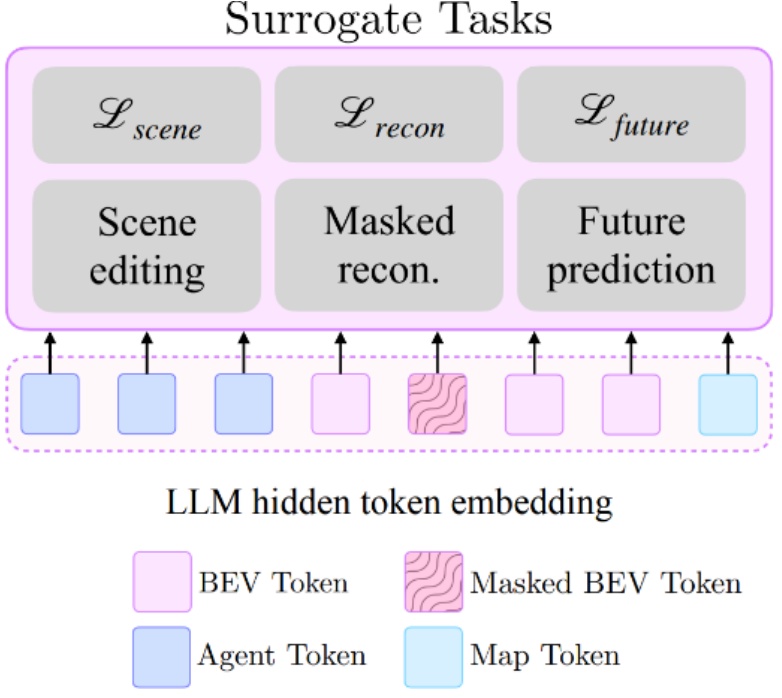}
    \vspace{-10pt}
    \caption{Overview of Surrogate tasks. Here hidden token embeddings are latent representations from the penultimate layer of LLM. These hidden token embeddings corresponding to \textbf{b}ird's-eye-view, \textbf{e}go, \textbf{a}gent, and \textbf{m}ap ($BEAM$) token embeddings are used as input as surrogate task decoder heads to perform masked reconstruction, future prediction and scene editing. }
    \label{fig:surrogate}
    \vspace{-10pt}
\end{wrapfigure}

An important component of training the MLLM branch of \ours~is the surrogate tasks module. In addition to being trained for planning and visual question answering, the MLLM is trained to perform the following surrogate tasks: masked reconstruction, future prediction, and scene editing. We design these tasks to enrich the \textbf{b}ird's-eye-view, \textbf{e}go, \textbf{a}gent, and \textbf{m}ap ($BEAM$) scene representations.  Surrogate tasks module takes hidden token embedding of penultimate layer of the LLM model. An illustration of the module can be seen in Figure ~\ref{fig:surrogate}. Each decoder head in surrogate module is consists of 3 Linear layers with a ReLU activation layer. Below, we provide some additional details on the scene editing tasks well as overviews on the other two tasks.

\noindent\textbf{Masked reconstruction.} Inspired by masked auto encoders \cite{he2022masked}, we formulate a masked reconstruction task of BEV token embeddings, to enrich the visual features learned by the scene encoder. The MLLM trained to infer masked regions based on the context provided by the visible BEV tokens and the rest of the multi-modal input. A reconstruction head takes the latent representations from the penultimate layer of LLM and predicts reconstructed BEV token embeddings $\hat{B}$. This decoder head is supervised using $\mathcal{L}_{recon}$ (refer equation 1 main paper). 

\noindent\textbf{Future prediction.} We formulate future prediction of BEV token embeddings, combined with VQAs to ground the scene encoder and MLLM model spatio-temporally, and learn robust spatio-temporal $BEAM$ scene representations. These spatio-temporal  $BEAM$ cues benefits planning performance (Table 4 of main paper). A future prediction head takes the latent representations from the penultimate layer of LLM and predicts future BEV token embedding at times $t+1, \: t+2$. We supervise these future BEV token embeddings  using $\mathcal{L}_{future}$ (refer equation 2 main paper).

\begin{figure*}[ht!]
    \centering
    \includegraphics[width=1\linewidth]{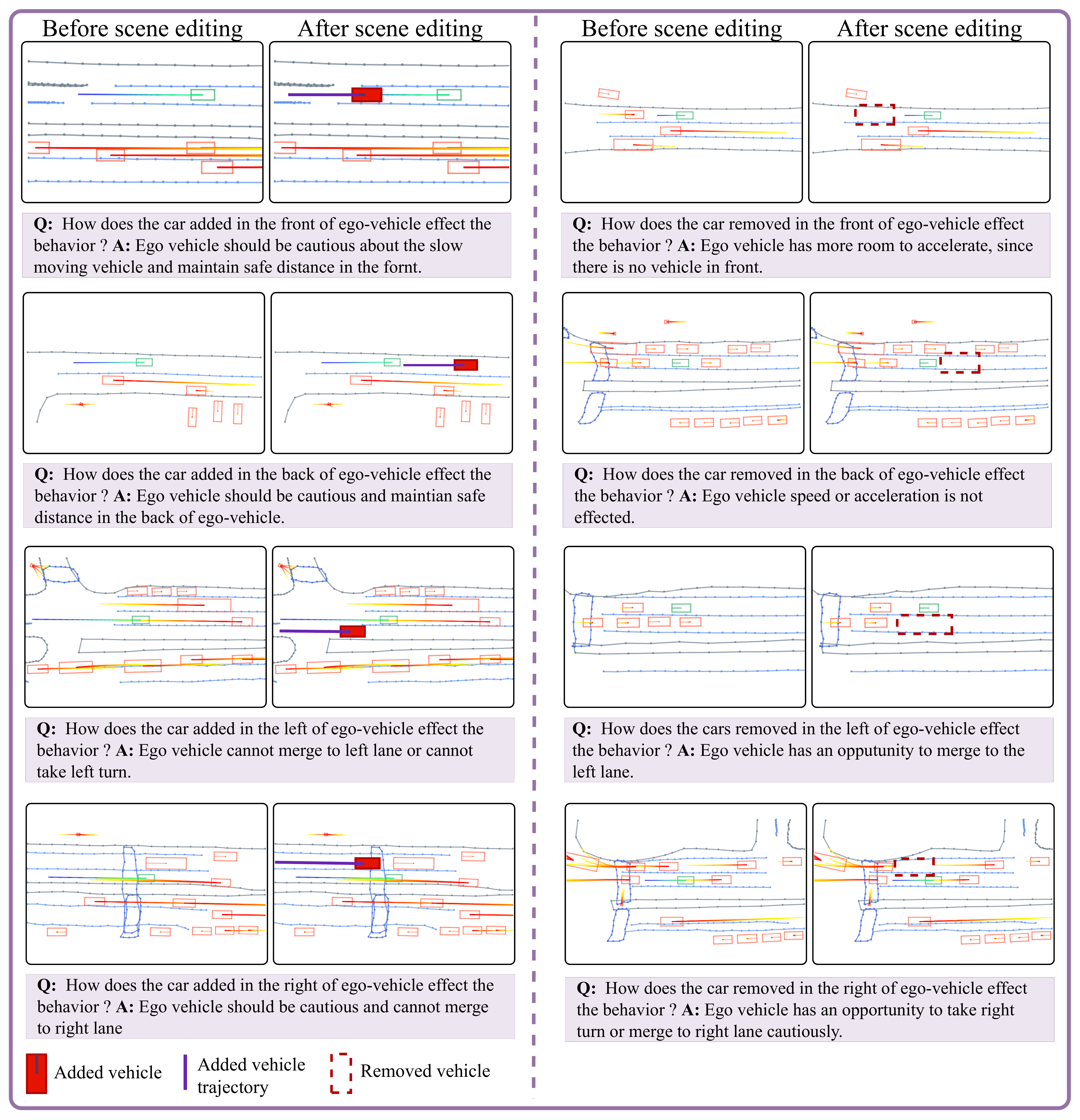}
    \caption{Examples of addition and deletion in scene editing.  In the left column, a car (solid red box) is added in the premises of the ego-vehicle (green box). In the right column, a car (dashed red box) is removed from the premises of the ego-vehicle. A corresponding question-answer pair is created to characterize the edit. }
    \label{fig:scene_edit_supp}
\end{figure*}

\noindent\textbf{Scene editing.} We propose a novel scene editing task in which we augment scenes by removing or adding new agents. Along with this, we construct a question-answer pair related to the edit. We show examples of this in Figure~\ref{fig:scene_edit_supp}. For scene addition, given the map constraints, predicted map, the ego bounding box location and the trajectories of predicted agents obtained from perception, and prediction tasks of scene encoding (refer \cite{hu2023planning,jiang2023vad}, using camera meta data we project all these to two-dimensional space as shown in \ref{fig:scene_edit_supp}. Using this, we identify possible locations in the premises of ego-vehicle location, where a new object can be added and randomly choose the location for the new agent, which is of size maximum $2\times$ of size of ego-vehicle. Given the location for the new agent and map constraints, we create a way-point trajectory for a new agent of category ``car" or ``truck". 
A new agent token embedding is then created using a linear layer. This new agent token embedding, the corresponding text prompt, and the rest of the $BEAM$ token embeddings are passed as input to the LLM. 
The hidden latent LLM features corresponding to agent token embeddings are then fed into a dedicated scene editing decoder head that performs way-point prediction of the ego vehicle. The output way-point prediction of the ego vehicle from scene editing head is supervised using $\mathcal{L}_{scene}$. $\mathcal{L}_{scene}$ is ego-agent collision constraint loss with updated agents incorporating either new added agent or removed agent in scene editing task. For ego-agent collision constraint loss refer \cite{jiang2023vad}.
The language prediction head performs question-answering on the new QA pair. This task thus contributes to the existing planning constraint loss and VQA loss of the MLLM. The QA pairs are generated according to a template in which the possible movements of the ego vehicle are described.


\section{Experimental Setup}
\label{sec:exp}
\subsection{Training details}
\label{sec:training}
We train \ours-VAD-tiny, \ours-VAD-Base, and \ours-UniAD, using training setups adapted from \cite{jiang2023vad} and \cite{hu2023planning}. Our training process follows a two-stage approach. First, we pre-train the vision planner only under the perception, prediction, and planning constraints for 60 epochs in order to learn informative latent scene representations. Second, we perform joint training of the vision planner and the MLLM for an additional 30 epochs, incorporating all proposed tasks and losses detailed in Section \ref{sec:method}. In the second stage, the language model of the MLLM is fine-tuned using LoRA \cite{hu2021lora}. Question-answer pairs from the augmented DriveLM dataset \cite{sima2023drivelm} are input along with the multi-view image sequence in the second stage. For each input sample, we randomly select one QA-pair from each category in every iteration to send as text prompt input to the network. This ensures diversity in questions while avoiding redundant visual inputs. For both stages, we employ the AdamW optimizer with a cosine annealing scheduler, a weight decay of 0.01, and an learning rate of $2 \times 10^{-4}$. In all our experiments we set the random masking ratio as $\left[0.2,0.4\right]$. 

\begin{wraptable}{r}{0.6\linewidth}
    \centering
    \vspace{-20pt}
    \caption{Comparison of L2 trajectory error and collision rate on nuScenes \cite{caesar2020nuscenes} using standardized evaluation \cite{weng2024drive}. Models are evaluated on the general validation split as well as a ``targeted'' split of challenging samples from \cite{weng2024drive}. The performance of the \ours~model variants are in shades of purple. We summarize results by averaging over at $t=\{1,2,3\}s$ as well as at all time steps. }
    \label{tab:supp}
    \resizebox{1\linewidth}{!}{
    \begin{tabular}{lccccccc}
    \hline
     \multicolumn{1}{c}{Method} & Using & \multicolumn{5}{c}{Traj L2 (m) $\downarrow$} & \multicolumn{1}{c}{Collision (\%) $\downarrow$} \\ \cline{3-8}
      & Ego status & 1s & 2s & 3s & \cellcolor{red!25}Ave$_{1,2,3s}$ & \cellcolor{red!25}Ave$_{all}$  &  \cellcolor{red!25}Ave$_{all}$ \\ \hline
    \rowcolor{gray!20} \multicolumn{8}{c}{\textbf{Full validation split}} \\ 
    UniAD\cite{hu2023planning} & \bccross & 0.48 & 0.89 & 1.47 & 0.95 & 0.83 &  0.40 \\
    PARA-Drive\cite{weng2024drive} & \bccross & 0.26 & 0.59 & 1.12 & 0.66 & 0.56 & 0.17 \\
    TOKEN\cite{tian2024tokenize} & \bccross & 0.26 & 0.70 & 1.46 & 0.81 & 0.68  & 0.15 \\
    \ccdb \ours (UniAD) &\ccdb \bccross &\ccdb 0.19 &\ccdb 0.50 &\ccdb 1.08 &\ccdb 0.59 &\ccdb 0.50 &\ccdb  0.06 \\ \hline
    \rowcolor{gray!20} \multicolumn{8}{c}{\textbf{Targeted validation split}} \\
    UniAD\cite{hu2023planning} & \bccross & 0.47 & 1.09 & 1.92 & 1.16 & 0.99 & 0.15 \\
    PARA-Drive\cite{weng2024drive} & \bccross & 0.38 & 0.97 & 1.88 & 1.08 & 0.91 & 0.14 \\
    
     \ccdb \ours (UniAD) &\ccdb  \bccross&\ccdb 0.30 &\ccdb 0.82 &\ccdb 1.63 &\ccdb 0.92 &\ccdb 0.77 &\ccdb 0.06 \\
    \hline
     \rowcolor{gray!20} \multicolumn{8}{c}{\textbf{3-point turn (zero-shot)}} \\
    UniAD\cite{hu2023planning} & \bccross & 0.68 & 1.55 & 2.90 & 1.71 & 1.43 & 0.00 \\
    PARA-Drive \cite{weng2024drive} & \bccross& 0.50 & 1.38 & 2.76 & 1.55 & 1.29 & 5.33 \\
    TOKEN \cite{tian2024tokenize} & \bccross &0.39 & 1.29 & 2.60 & 1.43 & 1.18 & 4.00 \\
    
     \ccdb \ours (UniAD) &\ccdb  \bccross&\ccdb 0.28 &\ccdb 0.94 &\ccdb 2.16 &\ccdb 1.13 &\ccdb 0.90 &\ccdb 0.00 \\
    \hline
     \rowcolor{gray!20} \multicolumn{8}{c}{\textbf{Resume from stop}} \\
    UniAD\cite{hu2023planning} & \bccross & 1.09 & 1.66 & 3.06 & 1.94 & 1.73 & 0.00 \\
    PARA-Drive & \bccross & 0.14 & 0.79 & 2.30 & 1.08 & 0.85 & 0.00 \\
    TOKEN & \bccross & 0.13 & 0.70 & 1.58 & 0.80 & 0.65 & 0.00 \\
     \ccdb \ours (UniAD) &\ccdb  \bccross&\ccdb 0.38 &\ccdb 0.83 &\ccdb 1.49 &\ccdb 0.90 &\ccdb 0.84 &\ccdb 0.00 \\
    \hline
     \rowcolor{gray!20} \multicolumn{8}{c}{\textbf{Overtake}} \\
    UniAD\cite{hu2023planning} & \bccross & 0.60 & 1.39 & 2.38 & 1.45 & 1.27 & 0.98 \\
    PARA-Drive & \bccross &  0.27 & 0.89 & 1.94 & 1.03 & 0.85 & 2.30 \\
     TOKEN & \bccross &  0.29 & 0.77 & 1.63 & 0.90 & 0.74 & 0.00 \\
     \ccdb \ours (UniAD) &\ccdb  \bccross&\ccdb 0.28 &\ccdb 0.75 &\ccdb 1.55 &\ccdb 0.86 &\ccdb 0.78 &\ccdb 0.41 \\
    \hline
    \end{tabular}
    }
    \vspace{-10pt}
\end{wraptable}

\subsection{Generation of text annotations}
\label{sec:data_gen}
We augment the existing text annotations of the Drive-LM \cite{sima2023drivelm} by generating question-answer (QA) pairs for samples in the nuScenes dataset \cite{caesar2020nuscenes}. First, we parse the numerical annotations of each object in the scene, such as the ego-vehicle and the surrounding objects. We denote each object as an agent and assign attributes such as the camera in which it is visible, the name of the object, and the vehicle speed. Additionally, we use rule-based algorithms to assign brief text descriptions of the future movement, the direction of movement relative to the ego-vehicle, the future speed, the type of interaction with the ego vehicle, and the probability of collision with the ego-vehicle. Using this annotation along with a few in-context examples, we prompt a Llama 3-70B model \cite{dubey2024llama} to generate 5 Drive-LM-like QA pairs for each category of question. The input system prompt can be seen in Figure \ref{fig:sys_prompt}. An example of a textual description of the numerical annotations can be seen in Figure \ref{fig:anno}. Examples of generated QA pairs can be seen in Figure \ref{fig:gen}.

\section{Additional Qualitative Results}
\label{sec:qual}
We present extensive qualitative results of \ours~. In Section \ref{sec:plan}, we present a comparison of planning performance of the vision-based planner on nuScenes. In Section \ref{sec:vqa}, we provide numerous visual question-answering results on various subsets of the nuScenes dataset.
\subsection{nuScenes planning}
\label{sec:plan}
\begin{figure*}
    \centering
    \includegraphics[width=0.8\linewidth]{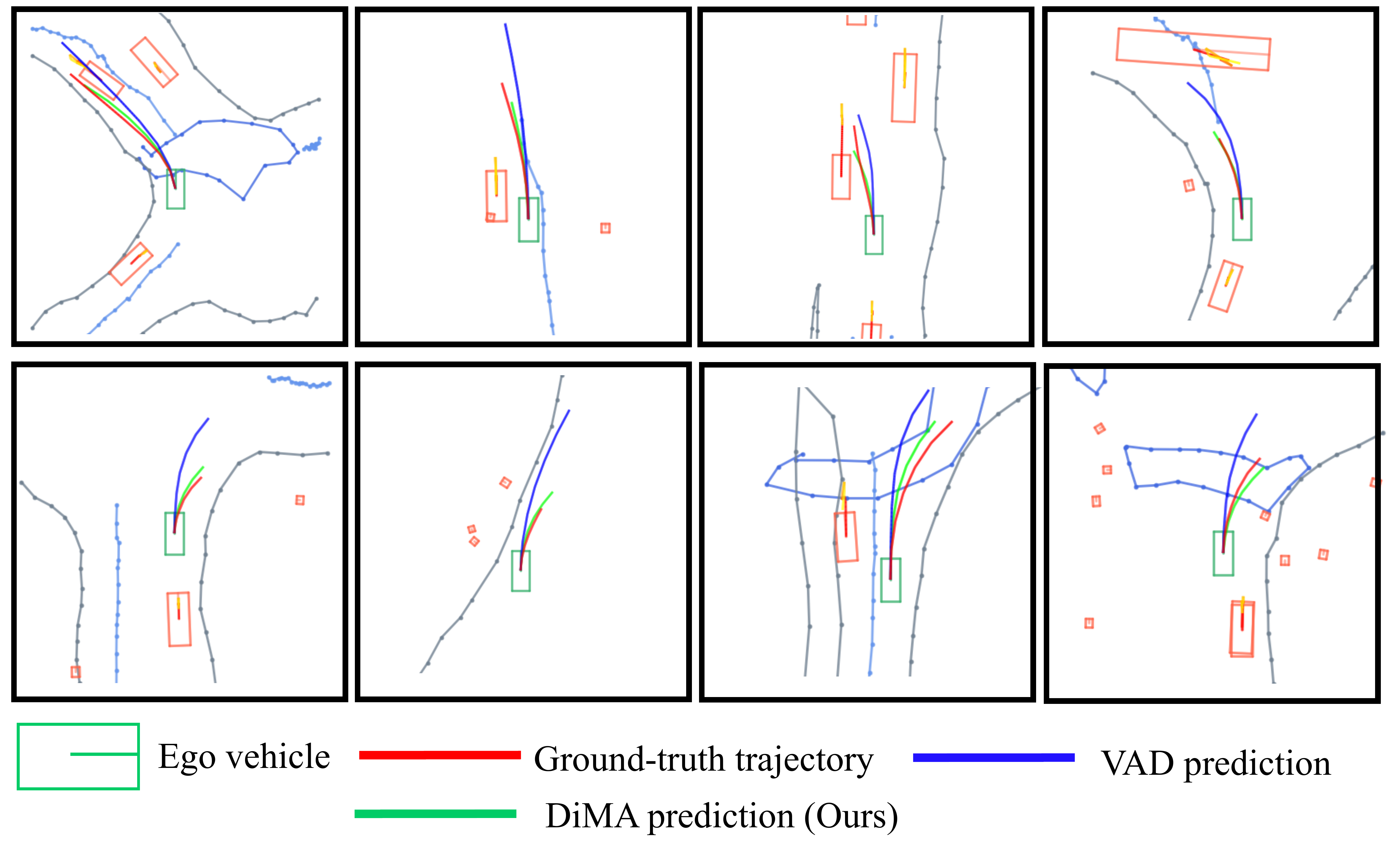}
    \caption{Visual comparison of the planning performance of \ours~(VAD-Tiny) with VAD-Tiny \cite{jiang2023vad}. Samples are from the ``targeted'' subset of the nuScenes validation split.}
    \label{fig:qual_plan}
\end{figure*}

We present qualitative planning results of \ours~compared to that of VAD in Figure \ref{fig:qual_plan}. We evaluate on difficult ``targeted'' samples from nuScenes. These are samples where the ego-vehicle is performing right and left turns. As seen in the figure, training with \ours~ensures safe trajectory prediction, avoiding collisions with vehicles when turning around corners (see row 1, columns 1 and 4) as well as avoiding lane departures (row 1 column 2). \ours~ also results in more precise turns (see row 2).

\begin{figure}[!htp]
    \centering

\begin{tcolorbox}[colback=cyan!10, colframe=cyan!50, boxrule=0.5mm, rounded corners, title={\textbf{\textcolor{black}{QA pair generation with Llama-3-70B \textbar{} System prompt}}}]

\begin{verbatim}
You are a system designed to generate high quality question answer pairs in the 
scenario of autonomous driving, from the point of view of an ego-vehicle 
viewing a 360 degree scene around you. Questions-answer pairs may be of three 
types : Perception, Prediction, and Planning. Perception questions relate to 
the nature of the agents/objects around the ego-vehicle. Planning questions 
relate to questions about the future actions of the ego-vehicle. Prediction 
questions are detailed questions about the  agents/objects around the ego 
vehicle. Here are some examples of each type of question.

Perception
	1. Q: 'What are objects to the front left of the ego car?'
	   A: 'There is one truck and one car to the front left of the ego car.' 
	2. Q: 'Are there moving pedestrians to the front right of the ego car?' 
	   A: 'Yes.' 

Prediction
	1. Q: 'Is <c1,CAM_FRONT> a traffic sign or a road barrier?' 
	   A: 'No' 
	2. Q: 'What object should the ego vehicle notice first when the ego 
    vehicle is getting to the next possible location? .... [TRUNCATED] 
	   A: 'Firstly, notice <c6,CAM_FRONT,1074.8,336.5>. It is a traffic sign, so 
       the ego vehicle should stop. Secondly, notice 
       <c1,CAM_FRONT,1413.3,534.2>. It is stationary, so .... [TRUNCATED] 

Planning 
	1. Q: 'What is the probability of colliding with <c1,CAM_FRONT> after the ego 
    vehicle steps on the brakes?' 
	   A: 'Low' 
	2. Q: 'In this scenario, what are safe actions to take for the ego vehicle?' 
	   A: 'Brake gently to a stop, slightly offset to the right.' 

For the given scene, generate five of each type of questions based on 
the attributes provided of the ego vehicle and each agent surrounding the ego
vehicle. Make sure that the answer is in the format 

    {
        'Perception':
        	   {'Q':'Question','A':'Answer'},
        'Prediction':
        	   {'Q':'Question','A':'Answer'},
        'Planning':
        	   {'Q':'Question','A':'Answer'}
    } 

Do not return any other text than the QA pairs in a correct python dictionary 
format. Make sure the answers are as descriptive as possible. Avoid one word
answers or yes/no questions.
\end{verbatim}

\end{tcolorbox}
    \caption{The system prompt given to Llama-3 to generate question-answer pairs.}
    \label{fig:sys_prompt}
\end{figure}

\begin{figure}
    \centering

\begin{tcolorbox}[colback=gray!10, colframe=gray!50, boxrule=0.5mm, rounded corners, title={\textbf{\textcolor{black}{QA pair generation with Llama-3-70B \textbar{} Sample numerical annotation}}}]

\begin{verbatim}
"token": "f9878012c3f6412184c294c13ba4bac3",
"scene_description": "Car overtaking, parking lot, pedestrians, pedestrian 
exiting car, objects on the ground",
"agent_attributes": {
    "c0": {
        "category": "truck",
        "speed": -0.00784316331860773,
        "assigned_cameras": [
            "CAM_BACK_LEFT"
        ],
        "future movement": "stopped",
        "future speed": "not moving",
        "direction": "towards from ego vehicle",
        "interaction_with_ego_vehicle_type": "none",
        "probability_of_collision_with_ego_vehicle": "low"
    },
    "c1": {
        "category": "pedestrian",
        "speed": 0.0,
        "assigned_cameras": [
            "CAM_FRONT",
            "CAM_FRONT_LEFT"
        ],
        "future movement": "stopped",
        "future speed": "not moving",
        "direction": "away from ego vehicle",
        "interaction_with_ego_vehicle_type": "none",
        "probability_of_collision_with_ego_vehicle": "low"
    },

    ... [TRUNCATED] ...
    
    "c24": {
        "category": "car",
        "speed": 0.0,
        "assigned_cameras": [
            "CAM_BACK"
        ],
        "future movement": "stopped",
        "future speed": "not moving",
        "direction": "away from ego vehicle",
        "interaction_with_ego_vehicle_type": "none",
        "probability_of_collision_with_ego_vehicle": "low"
    }

\end{verbatim}

\end{tcolorbox}
    \caption{An example of the text-description of the numerical annotations of a scene from nuScenes. This JSON file is created using rule-based algorithms. This is appended to the system prompt.}
    \label{fig:anno}
\end{figure}

\begin{figure}
    \centering

\begin{tcolorbox}[colback=gray!10, colframe=gray!50, boxrule=0.5mm, rounded corners, title={\textbf{\textcolor{black}{QA pair generation with Llama-3-70B \textbar{} Generated QA pairs}}}]

\begin{tcolorbox}[colback=green!10, colframe=green!50, boxrule=0.5mm, rounded corners, title={\textbf{\textcolor{black}{Perception}}}]

\begin{verbatim}
{
    "Q": "What are objects to the back right of the ego car?",
    "A": "There is one car and one bicycle to the back right of the ego 
         car."
},
{
    "Q": "Are there moving agents to the back right of the ego car?",
    "A": "Yes, there is one moving car to the back right of the ego car."
},
... [TRUNCATED] ...
\end{verbatim}
\end{tcolorbox}

\begin{tcolorbox}[colback=pink!10, colframe=pink!50, boxrule=0.5mm, rounded corners, title={\textbf{\textcolor{black}{Prediction}}}]

\begin{verbatim}
{
    "Q": "What are the future movements of the agents to the back right
         of the ego car?",
    "A": "The car will slightly steer to the right and the bicycle will 
         remain stopped."
},
{
    "Q": "What are the future speeds of the agents to the back right of 
         the ego car?",
    "A": "The car will be driving fast and the bicycle will not be 
         moving."
},
... [TRUNCATED] ...
\end{verbatim}
\end{tcolorbox}

\begin{tcolorbox}[colback=purple!10, colframe=purple!50, boxrule=0.5mm, rounded corners, title={\textbf{\textcolor{black}{Planning}}}]

\begin{verbatim}
{
    "Q": "What is the probability of colliding with the car after the 
         ego vehicle steps on the brakes?",
    "A": "Medium"
},
{
    "Q": "What actions taken by the ego vehicle can lead to a collision 
         with the bicycle?",
    "A": "No action taken by the ego vehicle will lead to a collision 
         with the bicycle."
},
... [TRUNCATED] ...
\end{verbatim}
\end{tcolorbox}

\begin{tcolorbox}[colback=orange!10, colframe=orange!50, boxrule=0.5mm, rounded corners, title={\textbf{\textcolor{black}{Behavior}}}]

\begin{verbatim}
    "Q": "Predict the behavior of the ego vehicle.",
    "A": "The ego vehicle is slighlty steering to the right. The ego 
         vehicle is moving at a moderate speed."
\end{verbatim}
\end{tcolorbox}

\end{tcolorbox}
    \caption{Some examples of generated QA pairs. The perception, prediction, and planning pairs are generated with Llama-3. The behavior QA is created using the future motion of the ego vehicle.}
    \label{fig:gen}
\end{figure}

\subsection{Visual question-answering}
\label{sec:vqa}
\begin{figure*}[t!]
    \centering
    \begin{subfigure}{0.49\textwidth}
        \includegraphics[width=\linewidth]{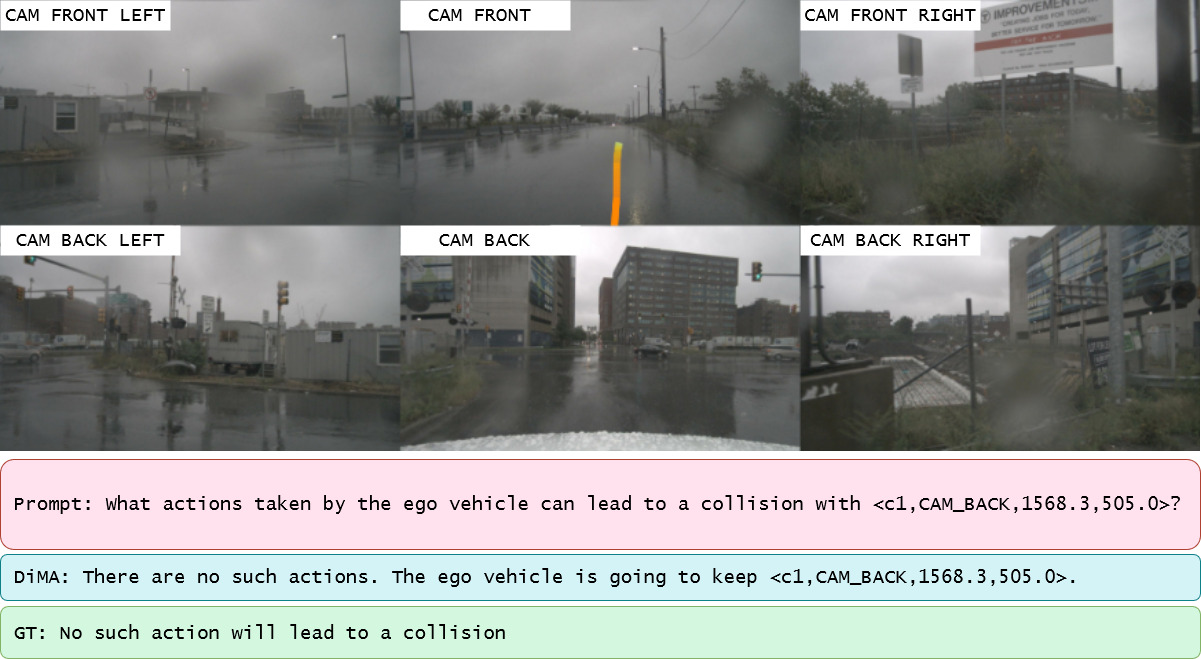}
    \end{subfigure}
    \hfill
    \begin{subfigure}{0.49\textwidth}
        \includegraphics[width=\linewidth]{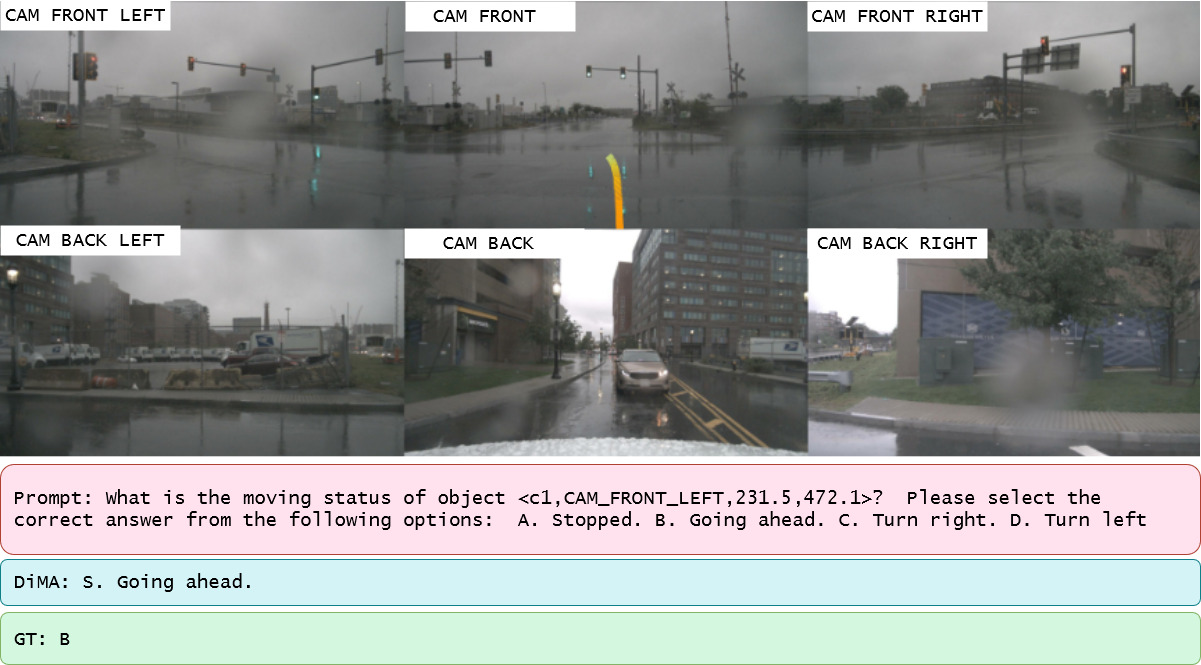}
    \end{subfigure}
    \vspace{-5pt}
    \vspace{8pt}

    \centering
    \begin{subfigure}{0.49\textwidth}
        \includegraphics[width=\linewidth]{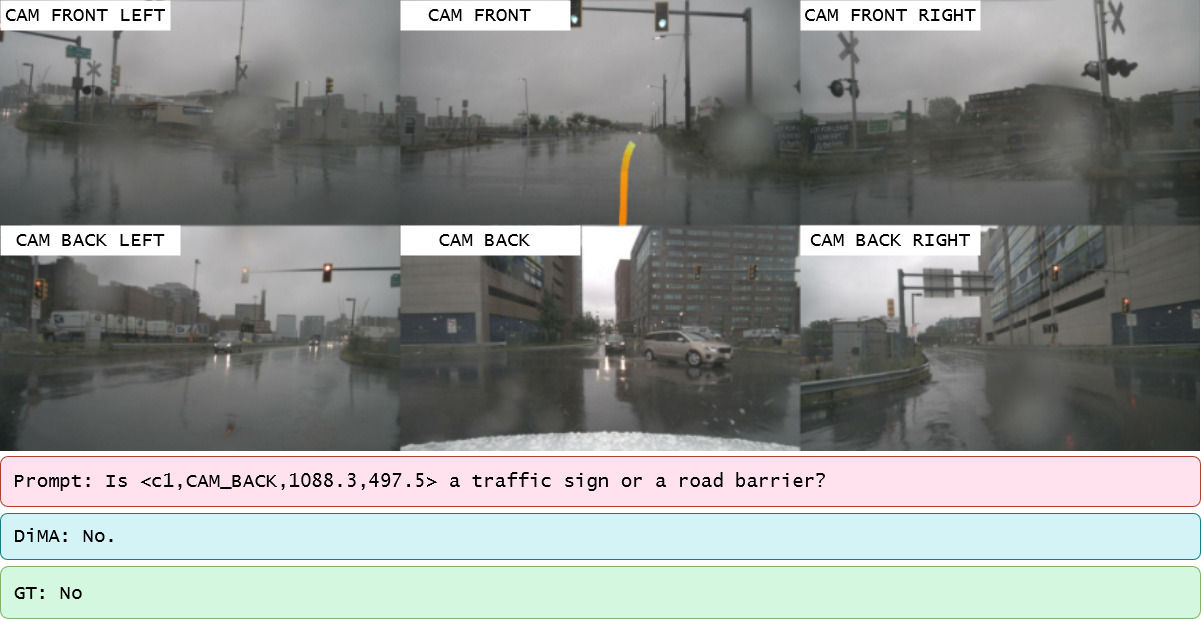}
    \end{subfigure}
    \hfill
    \begin{subfigure}{0.49\textwidth}
        \includegraphics[width=\linewidth]{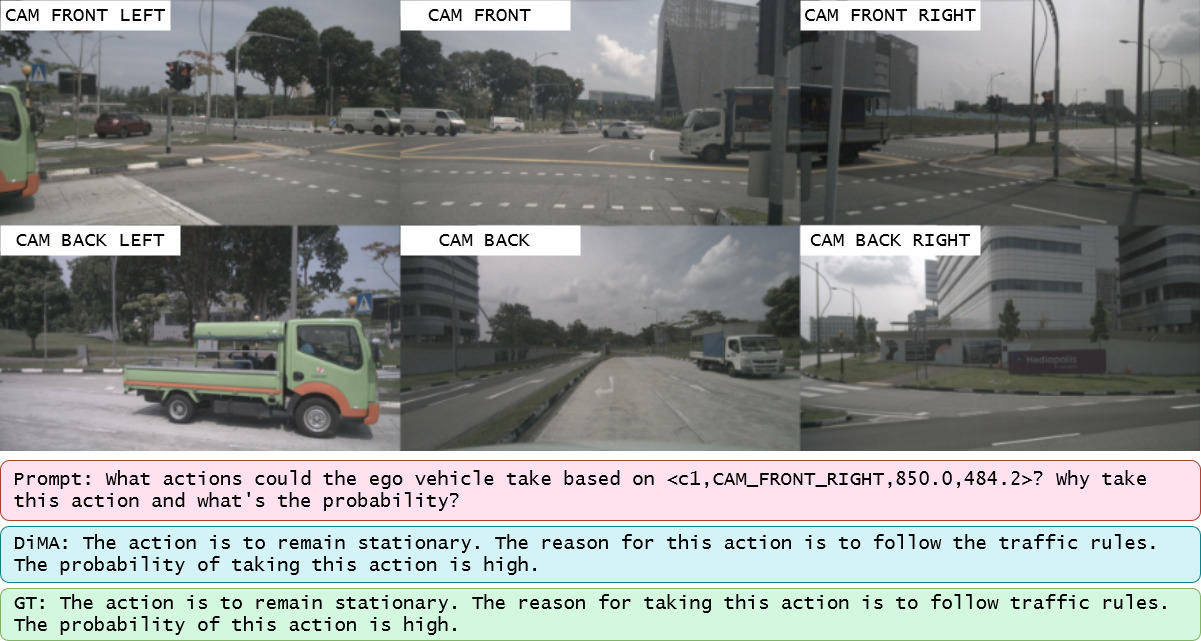}
    \end{subfigure}
    \vspace{-5pt}
    \vspace{8pt}

    \caption{Visualization of VQA and planning prediction by the MLLM branch of \ours-VAD-Tiny. We plot the predicted trajectory (orange-yellow) and show an example response of the language model branch to a question from the DriveLM test dataset \cite{sima2023drivelm}.}
    \label{fig:vqa_qual_supp}
    \vspace{-8pt}
\end{figure*}

\begin{figure*}[t!]
    \centering
    \begin{subfigure}{0.49\textwidth}
        \includegraphics[width=\linewidth]{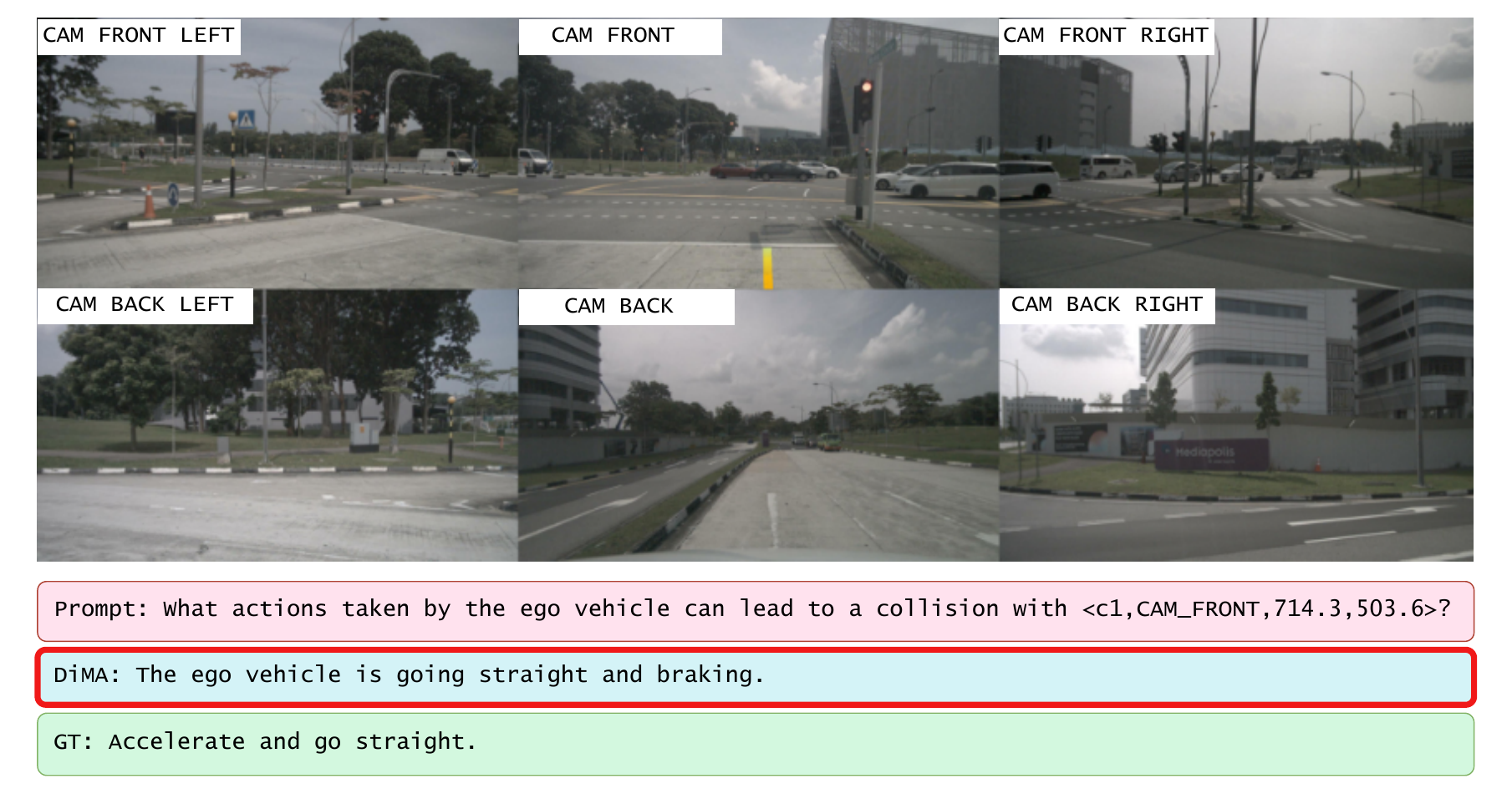}
    \end{subfigure}
    \hfill
    \begin{subfigure}{0.49\textwidth}
        \includegraphics[width=\linewidth]{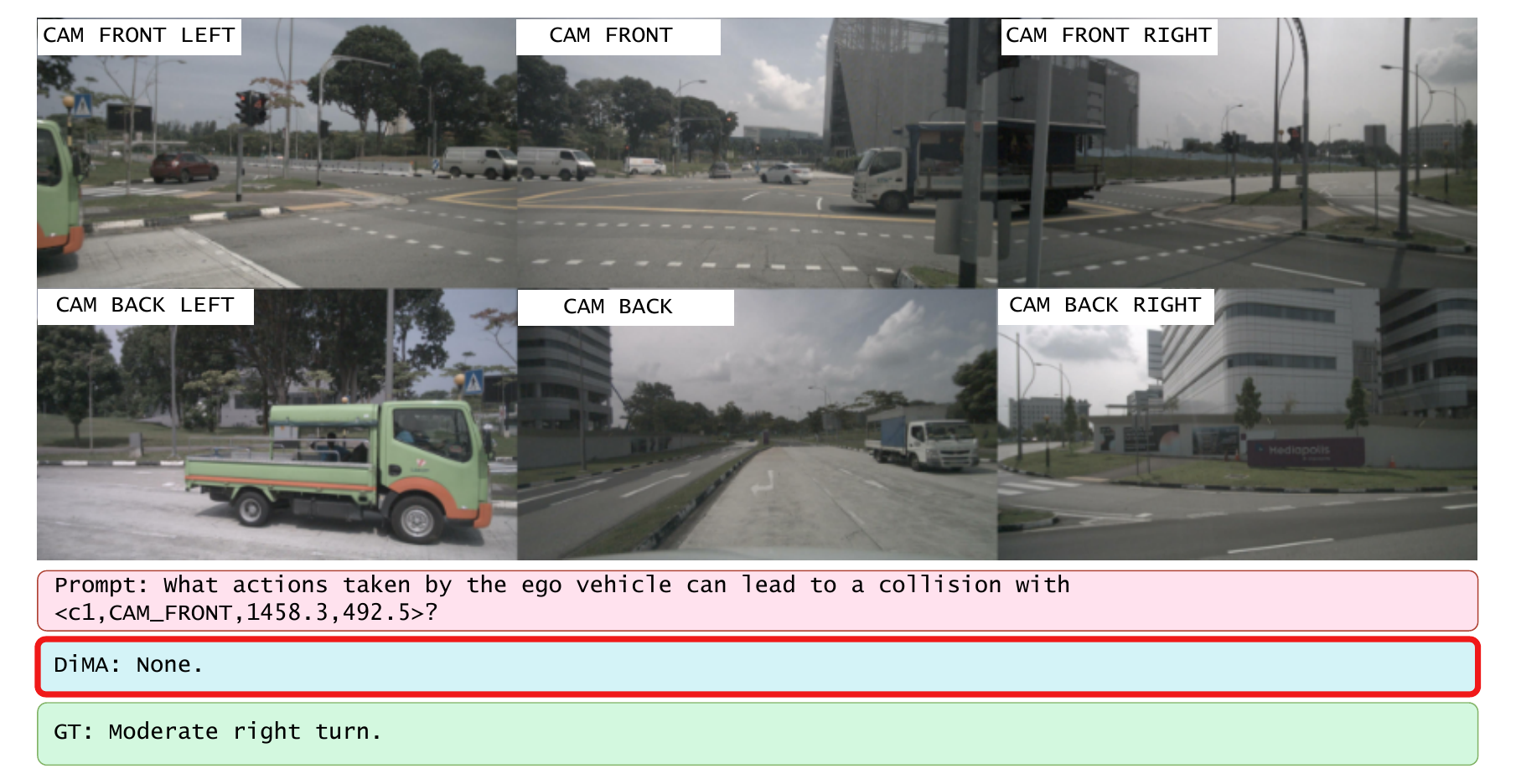}
    \end{subfigure}
    \vspace{-5pt}
    \caption{Visualization of failure cases of \ours-VAD-Tiny. We plot the predicted trajectory (orange-yellow) and show an example response of the language model branch to a question from the DriveLM test dataset \cite{sima2023drivelm}.}
    \label{fig:fail}
    \vspace{-8pt}
\end{figure*}

We present numerous qualitative examples of planning and VQA performance of the \ours~MLLM branch. For Drive-LM test samples that have ground-truth annotations, we compare the generated text response with the ground-truth answer in Figure \ref{fig:vqa_qual_supp}. We also plot the predicted future trajectory. In Figure \ref{fig:fail} we show two scenarios in which \ours~ provides incorrect VQA results.

\begin{figure*}[t!]
    \centering
    \begin{subfigure}{0.7\textwidth}
        \includegraphics[width=\linewidth]{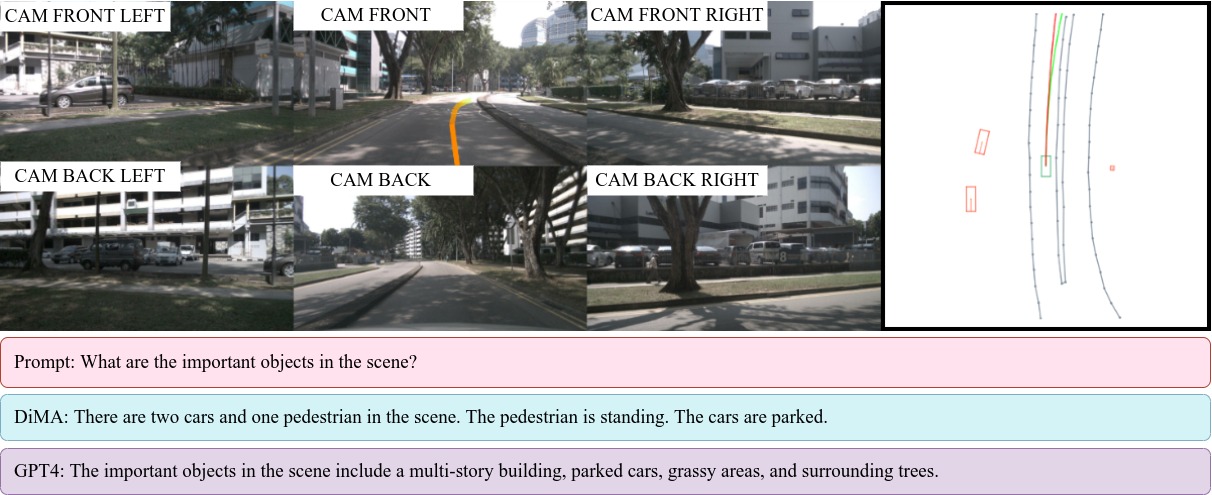}
    \end{subfigure}
    \begin{subfigure}{0.7\textwidth}
        \includegraphics[width=\linewidth]{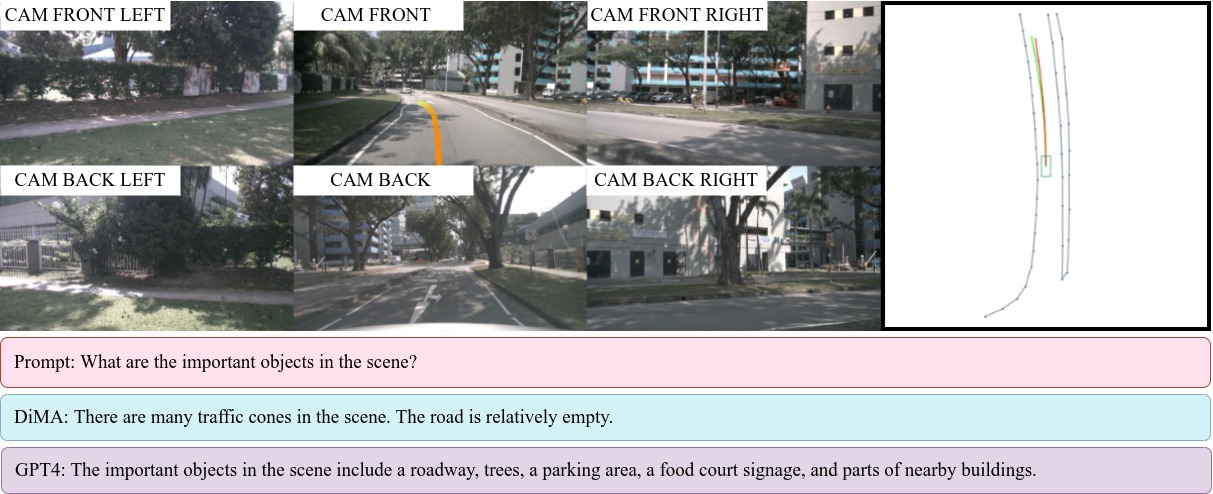}
    \end{subfigure}
    \vspace{-5pt}
    \begin{subfigure}{0.7\textwidth}
        \includegraphics[width=\linewidth]{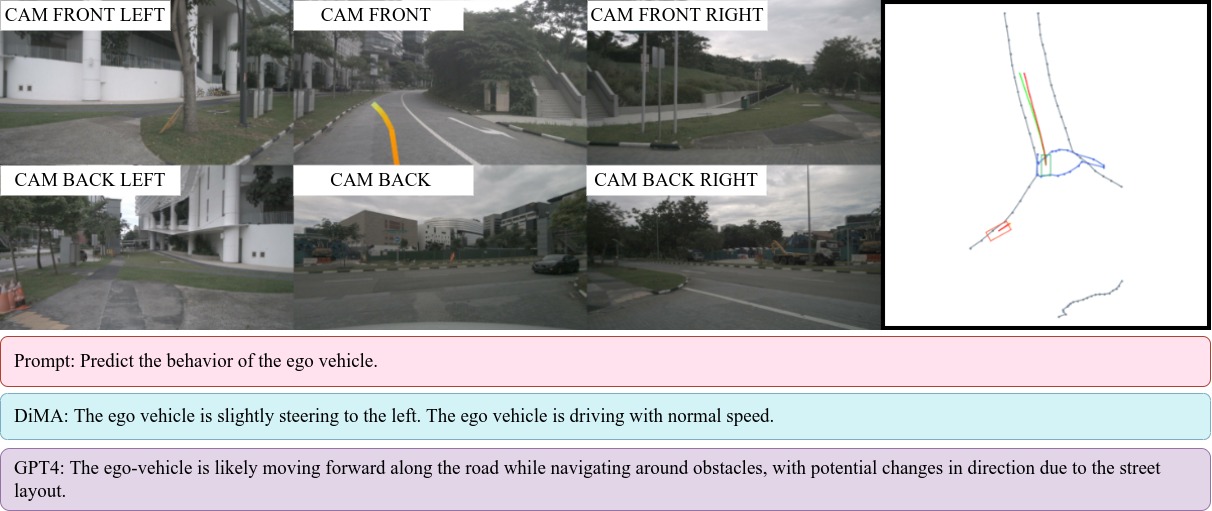}
    \end{subfigure}
    \vspace{-5pt}
    \begin{subfigure}{0.7\textwidth}
        \includegraphics[width=\linewidth]{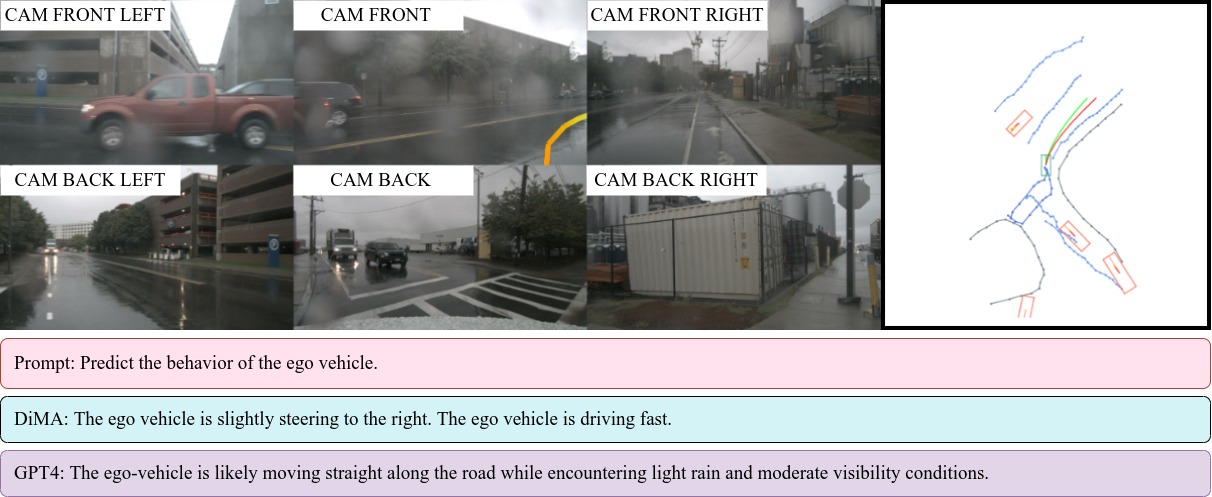}
    \end{subfigure}
    \vspace{-5pt}
    \caption{Visualization of visual question-answering on the targeted subset of the nuScenes dataset. On the image, we plot the predicted trajectory (orange-yellow) The red line is the ground-truth trajectory. In the diagram, the green line is the predicted trajectory.}
    \label{fig:gpt_qual_1}
    \vspace{-8pt}
\end{figure*}

\begin{figure*}[t!]
    \centering
    \begin{subfigure}{0.7\textwidth}
        \includegraphics[width=\linewidth]{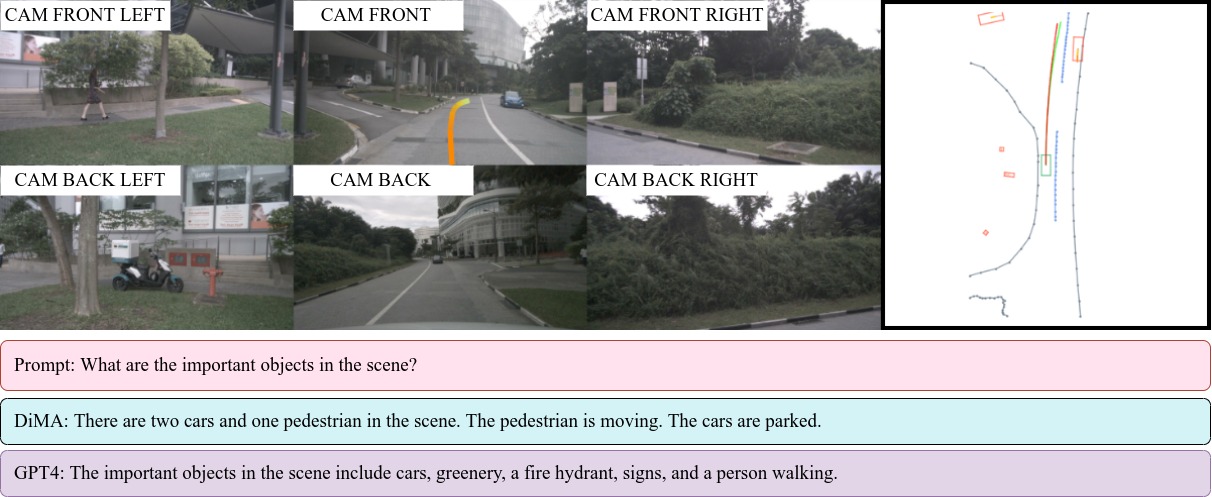}
    \end{subfigure}
    \begin{subfigure}{0.7\textwidth}
        \includegraphics[width=\linewidth]{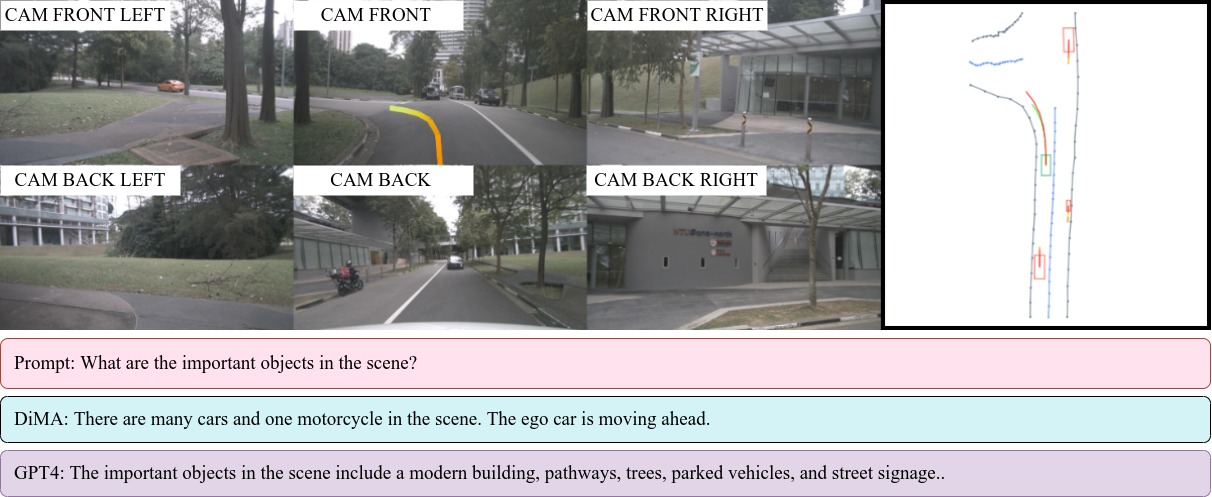}
    \end{subfigure}
    \vspace{-5pt}
    \begin{subfigure}{0.7\textwidth}
        \includegraphics[width=\linewidth]{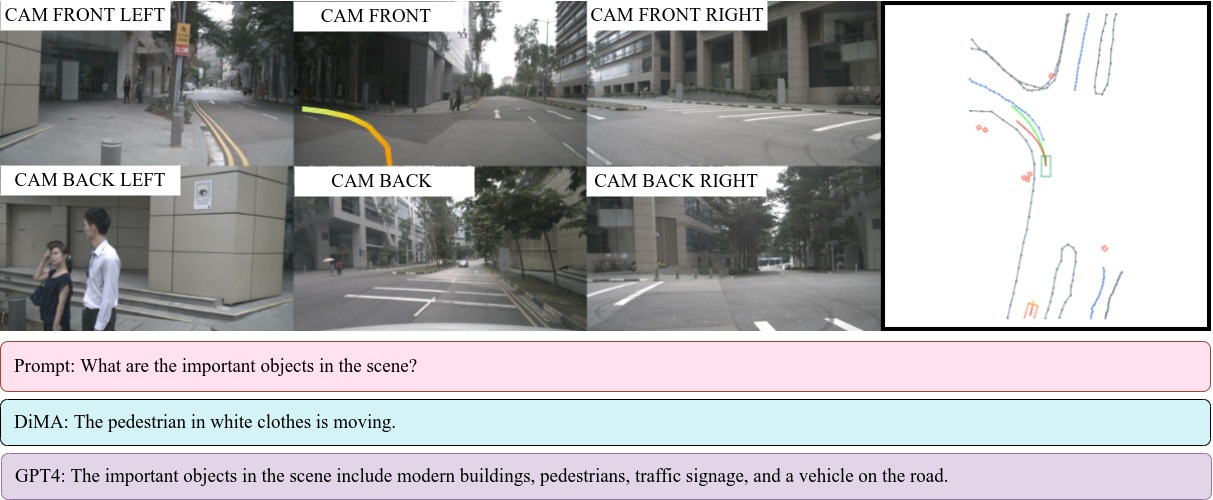}
    \end{subfigure}
    \vspace{-5pt}
    \begin{subfigure}{0.7\textwidth}
        \includegraphics[width=\linewidth]{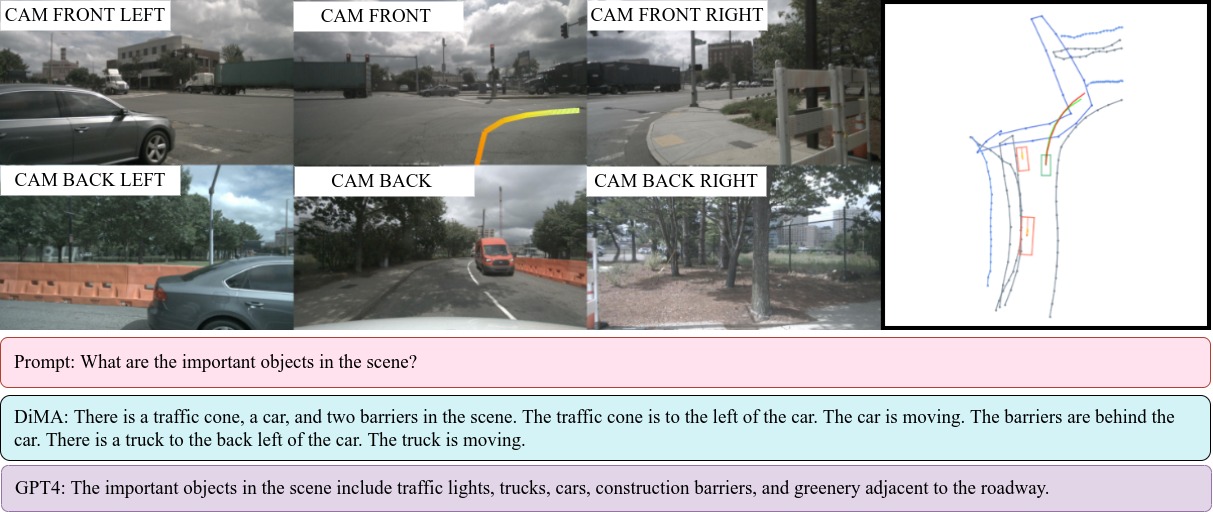}
    \end{subfigure}
    \vspace{-5pt}
    \caption{More visualization of visual question-answering on the targeted subset of the nuScenes dataset. On the image, we plot the predicted trajectory (orange-yellow) The red line is the ground-truth trajectory. In the diagram, the green line is the predicted trajectory.}
    \label{fig:gpt_qual_2}
    \vspace{-8pt}
\end{figure*}

\begin{figure*}[t!]
    \centering
    \begin{subfigure}{0.7\textwidth}
        \includegraphics[width=\linewidth]{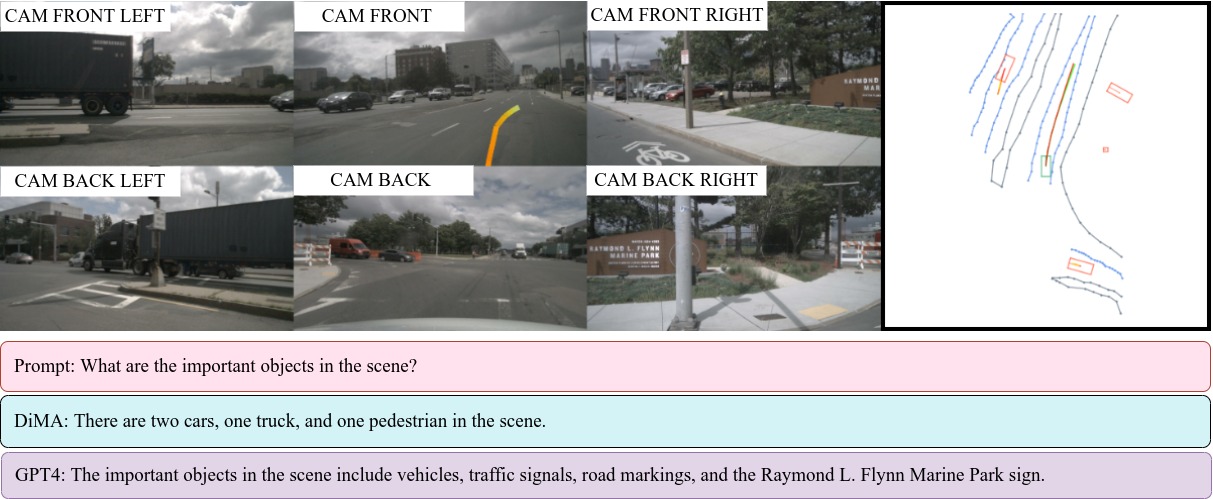}
    \end{subfigure}
    \begin{subfigure}{0.7\textwidth}
        \includegraphics[width=\linewidth]{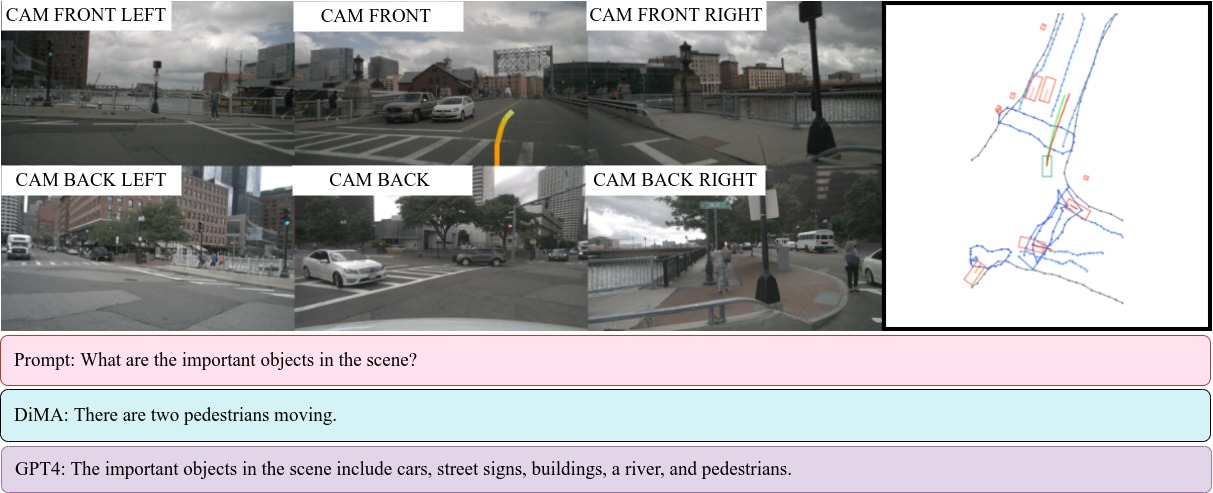}
    \end{subfigure}
    \vspace{-5pt}
    \begin{subfigure}{0.7\textwidth}
        \includegraphics[width=\linewidth]{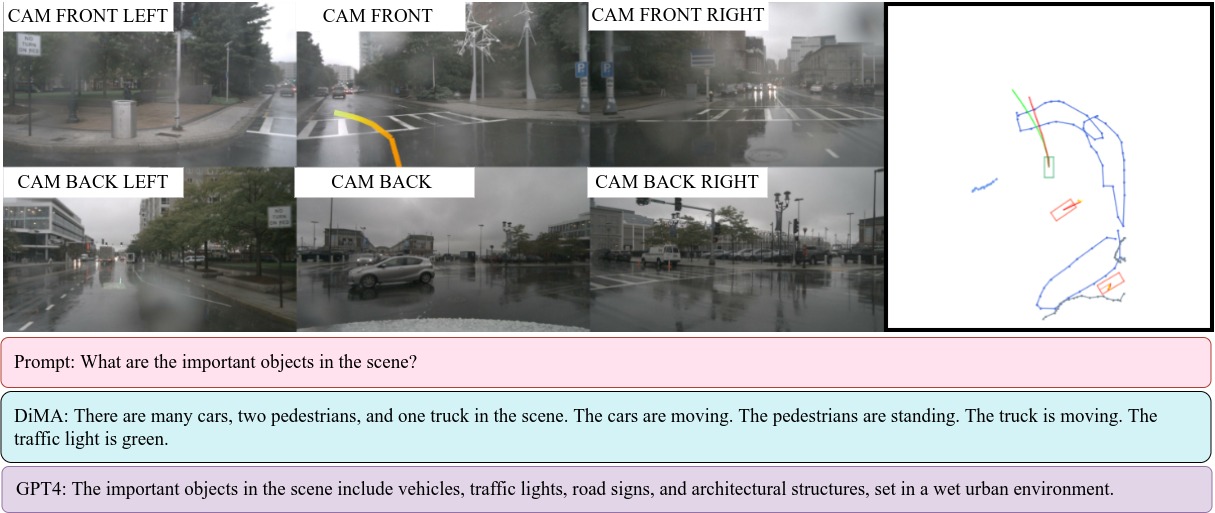}
    \end{subfigure}
    \vspace{-5pt}
    \begin{subfigure}{0.7\textwidth}
        \includegraphics[width=\linewidth]{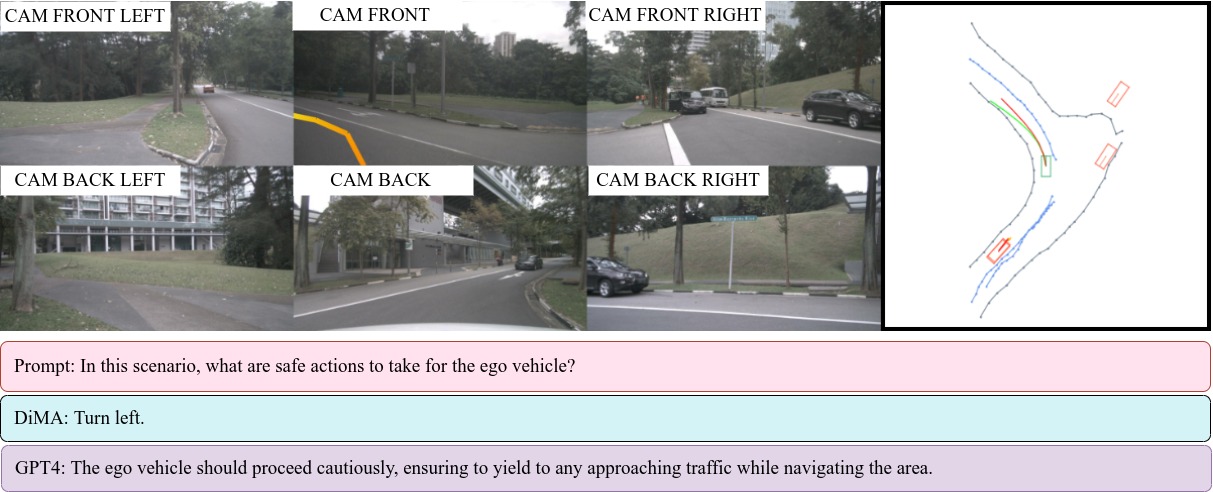}
    \end{subfigure}
    \vspace{-5pt}
    \caption{More visualization of visual question-answering on the targeted subset of the nuScenes dataset. On the image, we plot the predicted trajectory (orange-yellow) The red line is the ground-truth trajectory. In the diagram, the green line is the predicted trajectory.}
    \label{fig:gpt_qual_3}
    \vspace{-8pt}
\end{figure*}

For a more extensive qualitative analysis, we compare the performance of \ours~-MLLM with GPT-4 \cite{achiam2023gpt} in Figures \ref{fig:gpt_qual_1}, \ref{fig:gpt_qual_2}, and \ref{fig:gpt_qual_3}. We present common reasoning questions along with the \ours~-MLLM response and the GPT-4 response. The input to GPT-4 is the text prompt and a stitched image of the multi-view image set. We also show the planning performance plotted in the image as well as in a diagrammatic form on the right side of each row. As observed in these examples, \ours~is able to focus on objects important for navigation and planning. As can be seen in Figure \ref{fig:gpt_qual_1} row 4, our model correctly predicts the future right turn to be taken by the ego-vehicle, while GPT-4 suggests the ego-vehicle should move straight. A similar problem is observed in Figure \ref{fig:gpt_qual_3} row 4, where the prediction by GPT-4 is much more vague than that of \ours~.

\section{Additional Quantitative Results}
\label{sec:quant}
We present the performance of \ours-UniAD performance evaluated on the nuScenes dataset using standardized evaluation \cite{weng2024drive} in Table \ref{tab:supp}. We compare the performance of both \ours-UniAD and UniAD\cite{hu2023planning} on the general validation split as well as a ``targeted'' split of challenging samples from \cite{weng2024drive} and on long-tail scenarios. We observe consistent improvement across all metrics, resulting in significantly reduced L2 trajectory error and collision rate. This model version also out-performs state-of-the-art methods PARA-Drive \cite{weng2024drive} and TOKEN \cite{tian2024tokenize} in almost all cases.




\end{appendix}
\end{document}